\documentclass[10pt,journal]{IEEEtran}

\usepackage[pdftex]{graphicx}
\DeclareGraphicsExtensions{.pdf,.jpeg,.jpg}

\usepackage{multirow,setspace,verbatim,amsfonts,graphicx,amsmath,amsthm,amsbsy,amssymb,epsfig,url,cite}
\usepackage{amsthm}
\usepackage{gensymb}
\usepackage{graphicx}
\usepackage{hyperref}

\usepackage{amsfonts}
\usepackage{amsmath}
\usepackage{amssymb}
\usepackage{amsthm}
\usepackage{tikz,graphicx}
\usepackage{mathrsfs}

\usepackage{array}
\usepackage{booktabs}

\usepackage[algo2e,ruled]{algorithm2e}
\usepackage[noend]{algpseudocode}
\usepackage{booktabs, multirow, array}
\newcommand{\beq}{\begin{equation}}
\newcommand{\eeq}{\end{equation}}
\newcommand{\beqa}{\begin{eqnarray}}
\newcommand{\eeqa}{\end{eqnarray}}
\newcommand{\cb}{{\mathbf c}}
\newcommand{\db}{{\mathbf d}}

\newcommand{\gb}{{\mathbf g}}
\newcommand{\hb}{{\mathbf h}}

\newcommand{\tb}{{\mathbf t}}

\newcommand{\xb}{{\mathbf x}}
\newcommand{\yb}{{\mathbf y}}

\newcommand{\nub}{\boldsymbol{\nu}}

\newcommand{\alphab}{\boldsymbol{\alpha}}
\newcommand{\betab}{{\boldsymbol{\beta}}}

\newcommand{\Rd}{{\mathbb R}}
\newcommand{\Cd}{{\mathbb C}}

\newcommand{\hank}{\mathbb{H}}

\newcommand{\Mc}{{\mathcal M}}

\newcommand{\dmb}{\boldsymbol{\delta\mu}}
\newcommand{\psib}{\boldsymbol{\psi}}

\newcommand{\dm}{{\delta\mu}}

\newcommand{\add}[1] {\textcolor{black}{#1}} 

\DeclareMathAlphabet{\itbf}{OML}{cmm}{b}{it}

\def\bx{{{\bf x}}}

\definecolor{ao(english)}{rgb}{0.0, 0.5, 0.0}
\begin{document}

\title{Deep Learning Diffuse Optical Tomography}

\author{Jaejun~Yoo,~
        Sohail Sabir, Duchang Heo, Kee Hyun Kim, Abdul Wahab, Yoonseok Choi, Seul-I Lee, \\Eun Young Chae, Hak Hee Kim, Young Min Bae, Young-Wook Choi, Seungryong Cho,
        and~Jong~Chul~Ye$^{*}$,~\IEEEmembership{Senior Member,~IEEE}
\thanks{JY, SS, SC, and JCY are with Korea Advanced Institute of Science and Technology (KAIST), Daejeon 34141, Republic of Korea (e-mail: jaejun2004@gmail.com, \{sohail01, scho, jong.ye\}@kaist.ac.kr). JY is currently with the Clova AI Research, NAVER Corporation, Naver Green Factory, 6 Buljeong-ro, Bundang-gu,
13561, Republic of Korea (e-mail: jaejun.yoo@navercorp.com) and AW is currently with the NUTECH School of Applied Sciences and Humanities, National University of Technology, Islamabad 44000, Pakistan (e-mail: wahab@nutech.edu.pk).
DH, KHK, YMB, and YWC are with Korea Electrotechnology Research Institute (KERI), 111, Hanggaul-ro, Sangnok-gu, Ansan 15588, Republic of Korea (e-mail: \{dcheo, khkim1, kimbym, ywchoi\}@keri.re.kr).
YC is with Medical Research Institute, Gangneung Asan Hospital, 38, Bangdong-gil, Sacheon-myeon, Gangneung 210-711, Republic of Korea (e-mail: yschoi21rad@gmail.com).
SIL, EYC, and HHK are with Asan Medical Center, University of Ulsan College of Medicine, 88, Olympic-ro 43-gil, Songpa-gu, Seoul 05505, Republic of Korea (e-mail: si120919@gmail.com, chaeey@hanmail.net, hhkim@amc.seoul.kr ). This work is supported by the R\&D Convergence Program of NST (National Research Council of Science \& Technology) of Republic of Korea (Grant CAP-13-3-KERI).
}
}		

\maketitle

\begin{abstract}
Diffuse optical tomography (DOT) has been investigated as an alternative imaging modality for breast cancer detection thanks to its excellent contrast to hemoglobin oxidization level.
However, due to the complicated non-linear photon scattering physics and ill-posedness, the conventional reconstruction algorithms are sensitive to imaging parameters such as boundary conditions.
To address this, here we propose a novel deep learning approach that learns non-linear photon scattering physics and obtains an accurate three dimensional (3D) distribution of optical anomalies. In contrast to the traditional black-box deep learning approaches, our deep network is designed to invert the Lippman-Schwinger integral equation
using  the recent mathematical theory of deep convolutional framelets. As an example of clinical relevance, we applied the method to our prototype DOT system. We show that our deep neural network, trained with only simulation data, can accurately recover the location of anomalies within biomimetic phantoms and live animals without the use of an exogenous contrast agent.
\end{abstract}


\IEEEpeerreviewmaketitle

\section{Introduction}
\label{sec:introduction}

Deep learning approaches have demonstrated remarkable performance in many computer vision problems, such as image classification \cite{krizhevsky2012imagenet}. 
Inspired by these successes, recent years  have witnessed many innovative deep learning approaches  for various bio-medical image
reconstruction problems such as x-ray computed tomography,  photo-acoustics, ultrasound imaging, etc\cite{kang2017deep,antholzer2017deep,yoon2018efficient}.

Unlike these imaging applications where the measurement comes from linear operators,
there are other imaging modalities whose imaging physics should be described by complicated non-linear operators.
In particular, the diffuse optical tomography (DOT) is notorious due to severely non-linear and ill-posed operator originated from the diffusive
photon migration \cite{yodh1995spectroscopy,ntziachristos2000concurrent,boas2001imaging}. 
Although near-infrared (NIR) photons can penetrate several centimeters inside the tissue to allow non-invasive biomedical imaging,
the individual photons scatter many times and migrate along random paths before escaping from or being absorbed by the medium, which makes imaging task difficult. 
Mathematically, these imaging physics are  described by partial differential equations (PDE), and the goal is to recover constitutive parameters of the PDE from the scattered data measured at the boundary. This is called the \emph{inverse scattering problem}. 
Many dedicated mathematical and computational algorithms for the reconstruction of location and parameters of anomalies of different geometrical (cavities, cracks, and inclusions) and physical (acoustic, optical, and elastic) nature have been proposed over the past few decades \cite{markel2001inverse,jiang1996optical}. However, most of the classical techniques are suited for entire measurements or strong linearization assumptions \add{ under perfectly known boundary conditions}, which is usually not feasible in practice.

There are a few preliminary works that attempted to solve the inverse scattering problem using machine learning approaches \cite{kamilov2015learning,van2012method,sun2018efficient}.
For example, to obtain a non-linear inverse scattering solution for optical diffraction tomography, Kamilov et al. \cite{kamilov2015learning} proposed the so-called beam-propagation method
that computes the unknown photon flux using the back-propagation algorithm. 
The method by Broek and Koch \cite{van2012method} can be also considered an earlier version of the beam-propagation method using a neural network to calculate the dynamical scattering of fast electrons. {Sun et al. \cite{sun2018efficient} recently proposed a deep learning approach to invert  multiple scattering in optical diffraction tomography. 
However, these methods did not consider direct inversion of 3D distribution of anomalies.}

\add{Recently, Ye et al. \cite{ye2018deep,ye2019cnn} proposed a novel mathematical framework 
to understand deep learning approaches
in inverse problems. 
Rather than considering neural network as black-box, these frameworks lead to top-down design principle so that
 imaging application-specific knowledge can be used for neural network architecture design, even though
the specific type of nonlinearities, numbers of channels and filter sizes should be still tuned by trial and error. }


Specifically,  the non-linear mapping of Lippman-Schwinger type integral equations is fundamental in inverse scattering problems as shown in
\cite{lee2011compressive,lee2013joint},  so this perspective gives an idea how this imaging physics can be exploited in the design of the network.
In particular, our network is designed to invert the Lippman-Schwinger equation, but due to the ill-posed nature of the Lippman-Schwinger equation, we
impose an additional requirement that the output of the inverse mapping lies in a low-dimensional manifold.
Interestingly, by adding a fully connected layer at the first stage of the network followed by a CNN with an encoder-decoder structure,
this physical intuition is directly mapped to each layer of the convolutional neural network.

As a clinical relevance, we designed a DOT scanner as a part of simultaneous
X-ray {digital breast tomosynthesis} (DBT)  and DOT imaging system and 
applied the proposed network architecture as an inversion engine for optical imaging part. 
Although the network was trained only using the numerical data generated via the diffusion equation, extensive results using numerical- and real- biomimetic phantom as well as \emph{in vivo} animal experiments substantiate that the proposed method consistently outperforms the conventional methods. 

\section{Theory}
\label{sec:theory}
\subsection{Lippman-Schwinger integral equation}
\label{subsec:scatt}

In diffuse optical tomography
 \cite{yodh1995spectroscopy, ntziachristos2000concurrent, boas2001imaging,lee2011compressive,lee2013joint},
the basic assumption is that light scattering prevails over absorption. In this case, 
the propagation of light can be modeled by the diffusion equation. Let $\Omega$ be a domain filled with some turbid medium with $\partial \Omega$ as its boundary. In a highly scattering medium with low absorption, 
the total photon fluence rate $u(\xb)$ at position $\xb\in \Rd^3$   at the source modulation frequency $\omega$
can be modeled by the following frequency-domain diffusion equation:
\begin{eqnarray}\label{eq:DE}
\left\{ \begin{array}{ll}
\nabla \cdot D(\xb) \nabla u(\xb) - k^2(\xb) u(\xb) = -S(\xb), & \xb \in \Omega\\
u(\xb) + \ell\nub\cdot \nabla u(\xb) = 0, & \xb \in \partial \Omega
\end{array} \right.
\end{eqnarray}
where
 $\ell$ is an extrapolation length parameter related to the diffusion coefficient, dimension and reflection on the boundary; $\nub$ denotes a vector normal to the measurement surface,
 $\mu(\xb)$ and $D(\xb)$ are the absorption and diffusion coefficients, respectively;  $S(\xb)$ is the source intensity profile,
 and 
 the diffusive wave number $k$ is given by
$k^2(\xb):= \mu(\xb)-{i\omega}/{c_0} $
with $c_0$ denoting the speed of light in the medium.

In particular, our DOT system is mainly interested in the absorption parameter changes  due to the hemoglobin concentration changes:
\begin{eqnarray}\label{eq:mu}
\mu(\bx):= \mu_0(\bx)+\delta \mu(\bx) 
\end{eqnarray}
whereas $\mu_0$ denotes the known background absorption parameters and $\delta\mu$ refers its relative changes of the anomalies. Additionally,
the diffusion parameter $D$ is considered known. 
Then, the scattered fluence rate, $u_s(\xb):= u(\xb)- u_0(\xb)$, can be described by the so-called Lippman-Schwinger equation \cite{yodh1995spectroscopy, ntziachristos2000concurrent, boas2001imaging,lee2011compressive,lee2013joint}:
\begin{eqnarray}\label{eq:intDE}
u_s(\xb) 
&=& - \int_{\Omega} G_0(\xb,\yb) \delta \mu(\yb) u(\yb) d\yb,~
\end{eqnarray}
where  the background Green's function $G_0(\xb,\yb)$ satisfies
\begin{eqnarray}\label{eq:unpertrubedDE}
\left\{ \begin{array}{ll}
\left(\nabla \cdot D_0(\xb) \nabla - k_0^2(\xb)\right)G_0(\xb,\yb) = - \delta(\xb-\yb), & \xb \in \Omega\\
G_0(\xb,\yb) + \ell\nub\cdot \nabla G_0(\xb,\yb) = 0, & \xb \in \partial \Omega,
\end{array} \right. 
\end{eqnarray}
where $k_0 = \sqrt{\mu_0-i\omega/c_0}$ denotes the known background diffusive wave number, and 
 the incidence fluence $u_0(\xb)$ is given by
\begin{equation}\label{eq:U0}
u_0(\xb) = \int_{\Omega} G_0(\xb,\yb) S(\yb) d\yb .
\end{equation}

We assume that the absorption perturbation  is described by non-overlapping piecewise constant or spline approximation:
\begin{equation}\label{eq:Xmodel}
\delta\mu(\xb)= \sum\limits_{i=1}^{N} \delta \mu_i b_i(\xb),~~~
\end{equation}
where $b_i(\xb)$ denotes the  $i$-th basis function centered at $\xb_i\in \Omega$ and $\delta\mu_i$ is the corresponding coefficient.
Then, Eq.~\eqref{eq:intDE} can be represented by
\begin{eqnarray}\label{eq:intDEsampling}
u_s(\xb) 
       &=&  - \sum\limits_{i=1}^{N} G_0(\xb,\xb_i) u(\xb_i) \delta \mu_i 
\end{eqnarray}
Let $u_s^m(\xb)$ be the scattered photon fluence
at the $m$-th source intensity distribution given by
$$S_m(\xb)= S_0\delta(\xb-\tb_m),$$
where $\tb_m\in\Rd^3, m=1,\cdots, N_t$ denotes the point source location and $S_0$ is the source intensity.
We further assume the point detector at the detector location $\db_n\in \Rd^3,n=1,\cdots,N_d$.
Then, the measurement data can be described by 
 the multi-static data matrix:
 \begin{eqnarray}
 \gb&:=&\Mc[\dmb] \label{eq:msd_fwd}\\
 &:= &\begin{bmatrix} u_s^1(\db_1) &  u_s^2(\db_1) & \cdots & u_s^{N_t}(\db_1) \\ u_s^1(\db_2) &  u_s^2(\db_2) & \cdots & u_s^{N_t}(\db_2) \\ \vdots & \vdots & \ddots & \vdots \\ u_s^1(\db_{N_d}) 
&  u_s^2(\db_{N_d}) & \cdots & u_s^{N_t}(\db_{N_d}) 
 \end{bmatrix}   \in \Cd^{N_d\times N_t} \notag
 \label{eq:msd}
 \end{eqnarray}

The Born or Rytov approximation assumes that the optical perturbation $\delta\mu$ is sufficient small, so that 
the unknown fluence  $u(\xb)$ within the integral equation \eqref{eq:intDE}  can be approximated to the background
fluence rate, i.e. $u(\xb)\simeq u_0(\xb)$. This approximation, however, breaks down when the optical perturbation at the abnormalities are significant.
On the other hand,  the original form of Lippman-Schwinger equation in \eqref{eq:intDE} does not assume small perturbation so that
 the total optical photon density $u(\bx)$ also depends on the unknown perturbation.
This makes the inverse problem highly non-linear. Furthermore, due to the dissipative nature of the diffusive wave and a smaller number of measurements compared to the number of unknowns, reconstructing an image from the scattered optical measurement is a severely ill-posed problem \cite{arridge1999optical}. 
One could 
decouple the non-linear inverse problems from Lippman-Schwinger equation in two consecutive steps - joint sparse support recovery step 
and the linear reconstruction on the support- by taking advantage of the fact that the optical perturbation does not change position during multiple illuminations \cite{lee2011compressive,lee2013joint}.
In this paper, we further extend this idea  so that the optical anomalies can be directly recovered 
using a neural network.

\subsection{Neural network for inverting Lippman-Schwinger equation}\label{subsec:netArch}

The proposed neural network is designed based on the recent {\em deep convolutional framelets} for inverse problems \cite{ye2018deep},
so this section briefly reviews the theory.

For notational simplicity, we  assume that the absorption perturbation is one dimensional, but the extension to 3D is straightforward.
Specifically, let $\dmb=[\delta\mu[\xb_1],\cdots, \delta\mu[\xb_N]]^T\in \Rd^N$ and $\psib=[\psi[1],\cdots, \psi[d]]^T\in\Rd^d$  and
its reverse ordered version $\overline\psib[n] = \psib[-n]$, 
where the superscript $^T$ denotes the transpose operation, and $N$ and $d$ denote the number of voxel and the convolution
filter tap size, respectively.    
Then, the single-input single-output  convolution of an input $\dmb$ and a filter $\overline \psib$  can be represented in a matrix form:
\begin{eqnarray}\label{eq:SISO}
\yb = \dmb\circledast \overline\psib &=& \hank_d(\dmb) \psib \ ,
\end{eqnarray}
where  $\hank_d(\dmb)$ is a   Hankel matrix \cite{ye2018deep}:
 \begin{eqnarray} \label{eq:hank}
\hank_d(\dmb) =\left[
        \begin{array}{cccc}
        \dm[1]  &   \dm[2] & \cdots   &   \dm[d]   \\
       \dm[2]  &   \dm[3] & \cdots &     \dm[d+1] \\
           \vdots    & \vdots     &  \ddots    & \vdots    \\
              \dm[{N}]  &   \dm[1] & \cdots &   \dm[d-1] \\
        \end{array}
    \right] 
    \end{eqnarray}
%
Consider two matrices pairs ($\Phi, \tilde \Phi$) and ($\Psi$, $\tilde \Psi$) satisfying the conditions
\begin{align}
\tilde \Phi \Phi^\top = I_{{N\times N}}, \qquad \Psi \tilde \Psi^{\top} = P_{row}, \label{eq:id}
\end{align}
where 
$P_{row}$ represents a projection onto the row space of the Hankel matrix.
Then, we have
{
\begin{align}\label{eq:hankid}
\hank_d(\dmb) =\tilde\Phi\Phi^\top\hank_d(\dmb) \Psi \tilde \Psi^{\top} =\tilde\Phi C\tilde\Psi^{\top},
\end{align}}
with the coefficient matrix $C$ given by
\begin{align}\label{eq:coef0}
C = \Phi^\top\hank_d(\dmb) \Psi.
\end{align}
One of the most important discoveries in \cite{ye2018deep} is that an encoder-decoder structure convolution layer is emerged when the
high-dimensional Hankel matrix decomposition using
\eqref{eq:coef0} and \eqref{eq:hankid} is un-lifted to the orignal signal space.
Precisely, they are equivalent to the following paired encoder-decoder convolution structure:
\begin{eqnarray}
C &=& \Phi^\top \left( \dmb \circledast \alpha(\Psi)\right) \label{eq:enc} \\
&=& \Phi^\top \begin{bmatrix} \dmb \circledast \alphab_1 & \cdots & \dmb \circledast  \alphab_r \end{bmatrix}  \notag\\
\dmb &=& 
\left(\tilde\Phi C\right) \circledast \beta(\tilde \Psi) = \sum_{i=1}^r  \left(\tilde\Phi \cb_i \right) \circledast\betab_i  , \label{eq:dec} 
\end{eqnarray}
where
\eqref{eq:enc} corresponds to the single-input multi-output convolution,
and  \eqref{eq:dec} is the multi-input single-output convolution 
with the  encoder and decoder filters $\alpha(\Psi)=[\alphab_1\cdots\alphab_r]\in \Rd^{d\times r}$ and $\beta(\tilde\Psi)=[\betab_1\cdots\betab_r]\in \Rd^{d\times r}$ that are obtained by rearranging
the matrices $\Psi$ and $\tilde\Psi$, respectively \cite{ye2018deep}.
Note that the number of encoder and decoder filter channels are determined by  $r$ -
the rank of the Hankel matrix.

Now, we choose $\Phi=\tilde\Phi=I$ for simplicity so that the image resolution does not change during the filter process.
Then, by defining an inversion operator $\mathcal{T}:=\Mc^{-1}$ with respect to the forward operator $\Mc$ in \eqref{eq:msd_fwd} and substituting $\dmb=\mathcal{T} \gb$ in \eqref{eq:enc}, 
the encoder-decoder structure neural network can be re-written as
\begin{align}
C = \left( \mathcal{T} \gb \right) \circledast \alpha(\Psi),\qquad \dmb= \left(C\right) \circledast \beta(\tilde \Psi)
\label{eq:enc-dec}
\end{align}
where the coefficient $C=[\cb_1 \cdots \cb_r]$ at the decoder  can be replaced by $\hat C=[\hat \cb_1 \cdots \hat \cb_r]$ after removing noises using
the multi-input multi-output convolution:
\begin{eqnarray}\label{eq:MIMO}
\hat\cb_i = \sum_{j=1}^{r} \cb_j\circledast \hb_i^j,\quad i=1,\cdots, r
\end{eqnarray}
where 
$\hb_i^j \in \Rd^d$ denotes the length $d$- filter that convolves the $j$-th channel input to compute its contribution to 
the
$i$-th output channel.

\begin{figure}[!hbt]
\centerline{
\vspace*{-0.5cm}
\includegraphics[width=0.8\linewidth]{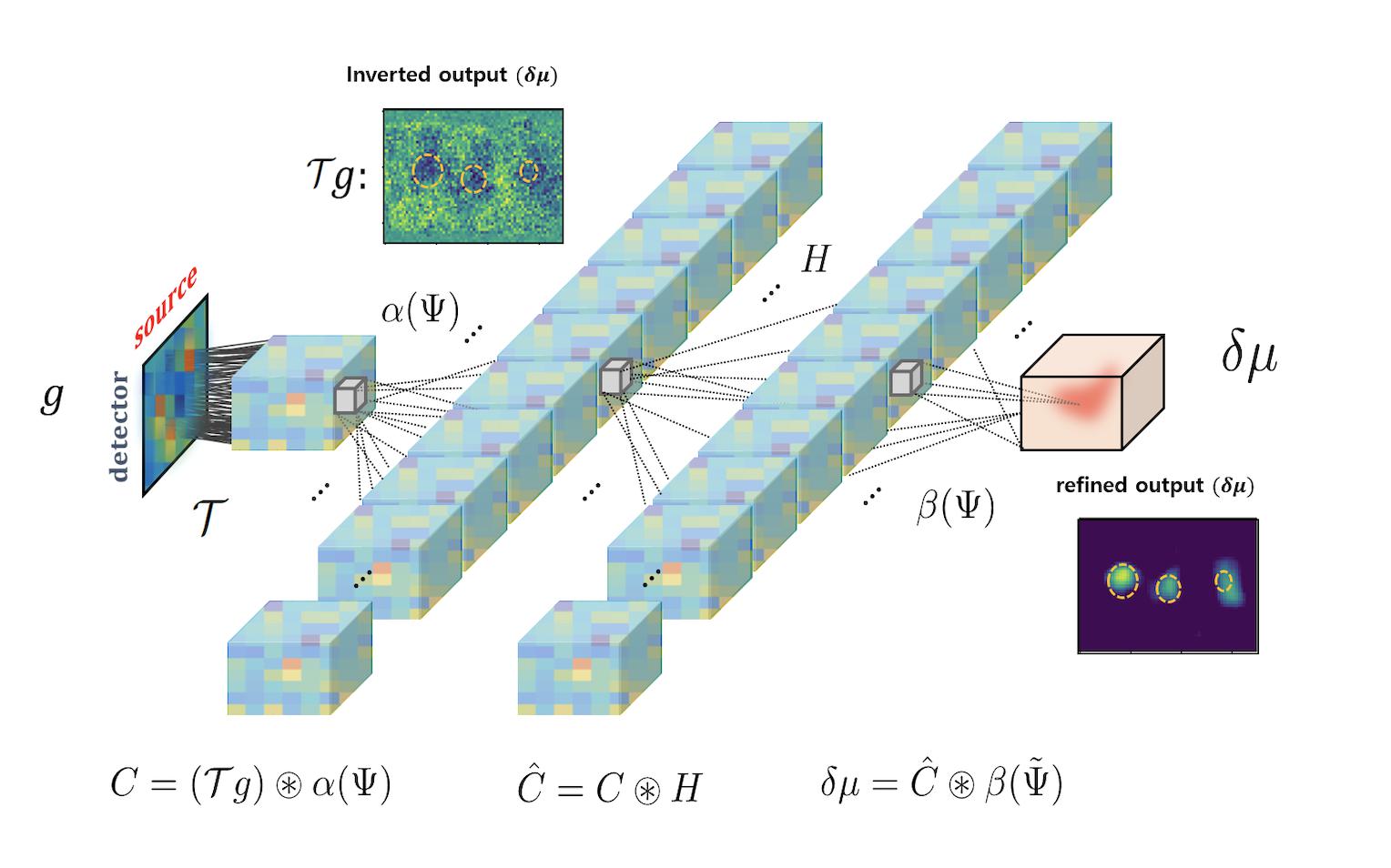}}
\caption{
Our neural network for inversion of the Lippman-Schwinger equation. 
 }
\label{fig:scc_net}
\end{figure}

The corresponding four-layer network structure is illustrated in Fig.~\ref{fig:scc_net}.
Here, the network consists of a single fully connected layer that approximates $\mathcal{T}$, two paired 3D-convolutional layers with filters $\alpha(\Psi)$ and $\beta(\tilde \Psi)$, and the intermediate 3D-convolutional filter $H=[\hb_1\cdots \hb_r]$ for additional filtering.
Then, the goal of our neural network is to learn the unknown fully connected layer mapping $\mathcal{T}$ and convolutional
filters, $\alpha(\Psi)$, $\beta(\tilde \Psi)$, and $H$, from the training data.

Here, it is important to note that  this parameter estimation may not provide the unique solution since
there are scale ambiguity in the estimation of these parameters.
Thus,  our fully connected layer $\mathcal{T}$ may be only a scaled and approximate version of the inverse operator $\Mc^{-1}$.
Moreover, although the convolutional framelet theory can give  a global perspective of network architecture,
the optimal hyper-parameters such as  
the number of the filter channel $r$, filter tap size  $d$, nonlinearity and the number of intermediate filter steps should be
found by trial and error.
\add{In particular,
our subsequent work in \cite{ye2019cnn} shows that the nonlinearities makes the  aforementioned
decomposition structure
 automatically adapted to different inputs, making the neural network generalizable.
}

%

Despite of these limitations,
 the proposed deep network has many advantages. First, the inversion of the Lippman-Schwinger equation is fully data-driven such that we do not need any explicit modeling of the acquisition system and boundary conditions. Second, 
 the manifold dimensionality of the optical parameter distribution is directly controlled by the number of  convolutional layers, $r$,  which is also  related to the redundancies of the $\dmb$.
In particular, specific manifold structure is  learned from the training data in the form of the convolution filters, which makes the algorithm
less affected by the measurement data deviation from the analytic diffusion model.

\section{Methods}\label{sec:methods}

\begin{figure}[!b]
\centering
\vspace*{-0.5cm}
\includegraphics[width=1\linewidth]{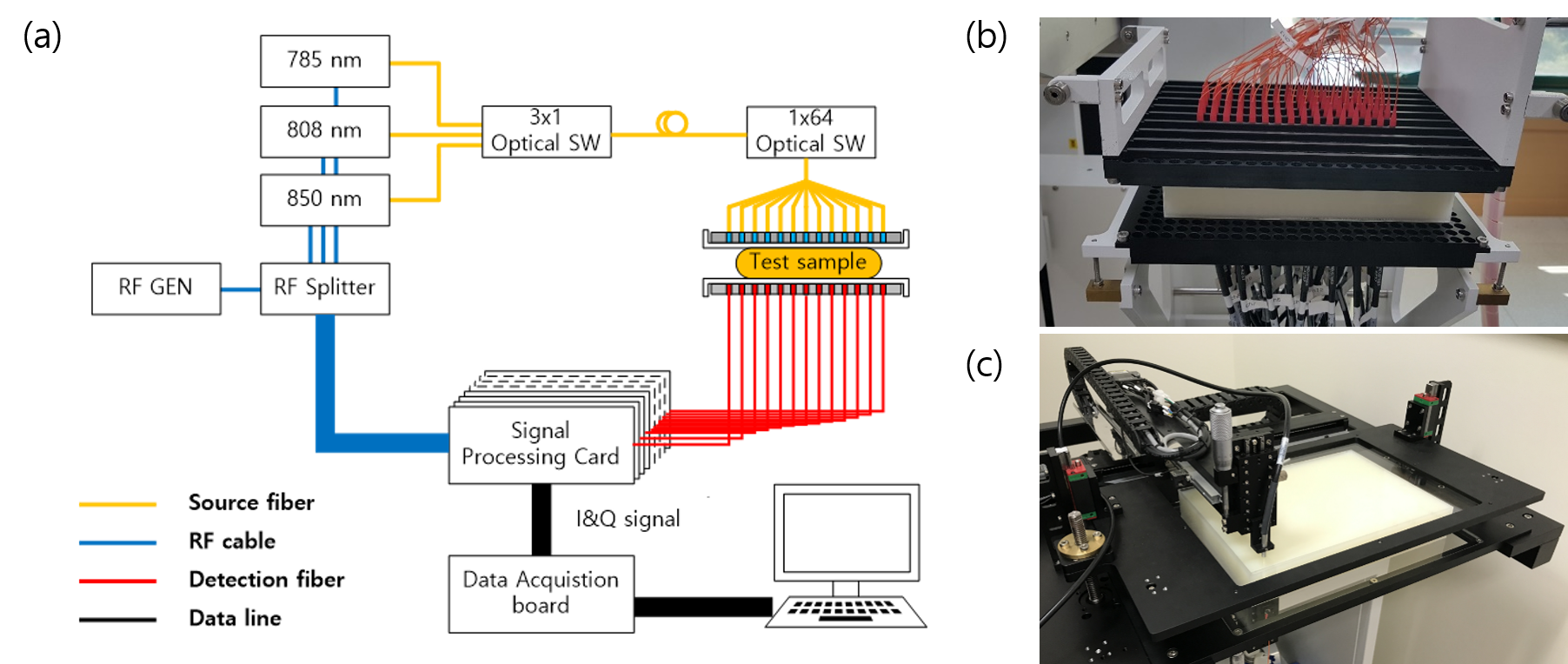}
\caption{
{Single- and multi-channel DOT systems used in our experiments \cite{heo2017spie,heo2017kps}. 
(a) Schematic illustration of the DOT system configuration. The light source has three fiber pigtailed laser diode modules with 785 nm, 808nm, and 850 nm. 70 MHz RF signal is simultaneously applied to these light sources using bias-T, RF splitter, and RF AMP. Two optical switches are used to deliver light to 64 specific positions in the source probe. During optical switching time, one-tone modulation light photons reach 40 detection fiber ends after passing an optical phantom and are detected simultaneously by 40 avalanche photodiodes (APD) installed in the home-made signal processing card. Real snapshots of our (b) multi-channel DOT, and (c) single-channel DOT systems. } }
\label{fig:hardware}
\end{figure}

Fig.~\ref{fig:hardware}(a) shows the schematics of our frequency domain DOT system. The DOT system has been developed at the Korea Electrotechnology Research Institute (KERI) to improve the diagnostic accuracy of the digital breast tomosynthesis (DBT) system  for joint breast cancer diagnosis \cite{heo2017spie,heo2017kps}. 
The multi-channel DOT system (Fig.~\ref{fig:hardware}(b))  consists of four parts: light source, optical detector, optical probe, and data acquisition and controller. The light source has three fiber pigtailed laser diode modules of 785 nm, 808 nm, and 850 nm. 70 MHz RF signal is simultaneously applied to these light sources using bias-T, RF splitter, and RF AMP. Two optical switches are used to deliver light to 64 specific positions in the source probe. During the optical switching time, one-tone modulation light photons reach 40 detection fiber ends after passing an optical phantom and are detected simultaneously by 40 avalanche photodiodes (APD) installed in the home-made signal processing card. The DOT system uses an In-phase(I) and Quadrature(Q) demodulator to get amplitude and phase of the signal in the signal processing card. The 40 IQ signal pairs are simultaneously acquired using data acquisition boards. The data acquisition time for all measurements took about 30 seconds.
For the purpose of preclinical tests, a single-channel system (Fig.~\ref{fig:hardware}(c)) has been installed at Asan Medical Center (AMC) \cite{choi2017kps,choi2017kiee}. The overall configuration of the single-channel DOT system is same as the multi-channel DOT system except for the number of light sources and optical detectors used. Here, it has only one source fiber and one optical detector. These fibers are installed in a motorized probe stage which is driven by highly precise stepping motors and control modules. Fig.~\ref{fig:hardware}(c) shows the probe stage. An operator can freely set up the scanning position and sequence according to the region of interest. The system includes only one optical switch for the selection of three wavelengths. 

\subsection{DOT Hardware System}\label{subsec:sys}

%

\begin{table*}[!hbt]
\centering
\caption{
{Network architecture specifications. Here, \emph{\#MEAS} is the number of filtered measurement pairs (Biomimetic: $\#MEAS=538$, {Numerical \& Breast-mimetic}: $\#MEAS=466$, Mouse (normal): $\#MEAS=470$, Mouse (with tumor): $\#MEAS=1533$).}}
\label{tb:net}
\resizebox{0.6\paperwidth}{!}{
\begin{tabular}{|c|c|c|c|c|c|c|c|c|c|}
\hline
\multirow{2}{*}{\textbf{Type}} & \multicolumn{3}{c|}{\textbf{Biomimetic}} & \multicolumn{3}{c|}{\textbf{{Numerical \& Breast-mimetic}}} & \multicolumn{3}{c|}{\textbf{Animal}}\\ \cline{2-10} 
& \textbf{\begin{tabular}[c]{@{}c@{}}patch size\\ /stride\end{tabular}} & \textbf{\begin{tabular}[c]{@{}c@{}}output\\ size\end{tabular}} & \textbf{depth} & \textbf{\begin{tabular}[c]{@{}c@{}}patch size\\ /stride\end{tabular}} & \textbf{\begin{tabular}[c]{@{}c@{}}output\\ size\end{tabular}} & \textbf{depth} & \textbf{\begin{tabular}[c]{@{}c@{}}patch size\\ /stride\end{tabular}} & \textbf{\begin{tabular}[c]{@{}c@{}}output\\ size\end{tabular}} & \textbf{depth}\\ \hline
\textbf{Network input} & - & $1\times \mbox{\emph{\#MEAS}}$ & - & - & $1\times \mbox{\emph{\#MEAS}}$ & - & - & $1\times \mbox{\emph{\#MEAS}}$ & - \\ \hline
\textbf{Fully connected} & - & $32 \times 64\times 20\times 1$ & - & - & $48 \times 70\times 16 \times 1$ & - & - & $32 \times 32\times 12 \times 2$ & - \\ \hline
\textbf{3D convolution} & $3\times 3\times 3$/1 & $32 \times 64\times 20\times 16$ & 16 & $3\times 3\times 3$/1 & $48 \times 70\times 16\times 64$ & 64& $3\times 3\times 3$/1 & $32 \times 32\times 12\times 128$ & 128 \\ \hline
\textbf{3D convolution} & $3\times 3\times 3$/1 & $32 \times 64\times 20\times 16$ & 16 & $3\times 3\times 3$/1 & $48 \times 70\times 16\times 64$ & 64 &$3\times 3\times 3$/1 & $32 \times 32\times 12\times 128$ & 128 \\ \hline
\textbf{3D convolution} & $3\times 3\times 3$/1 & $32 \times 64\times 20\times 1$ & 1 & $3\times 3\times 3$/1 & $48 \times 70\times 16\times 1$ & 1 &$3\times 3\times 3$/1 & $32 \times 32\times 12\times 1$ & 1 \\ \hline
\end{tabular}
}
\end{table*}

\begin{table*}[!hbt]
\footnotesize
\centering
\caption{Specification of FEM mesh for test data generation.}
\resizebox{0.6\paperwidth}{!}{
\begin{tabular}{|c|c|c|c|c|c|c|c|}
\hline \noalign{\hrule height 0.5pt}
\multirow{2}{*}{\textbf{}} & \multirow{2}{*}{\textbf{\# of sources}} & \multirow{2}{*}{\textbf{\# of detectors}} & \multicolumn{2}{c|}{\textbf{FEM mesh}} & \multicolumn{2}{c|}{\textbf{\begin{tabular}[c]{@{}c@{}}Optical parameters\\ Background ($mm^{-1}$)\end{tabular}}} & \multirow{2}{*}{ \textbf{\begin{tabular}[c]{@{}c@{}} \# of voxels per xyz dimensions\\(resolution$=2.5~mm$)\end{tabular}}}\\ \cline{4-7} 
& & & \textbf{nodes} & \textbf{elements} & \multicolumn{1}{c|}{\textbf{$\mu$ (absorption)}} & \textbf{$\zeta$ (scattering)} & \\ \hline
\textbf{Biomimetic phantom} & $4 \times 16$ & $5 \times 8$ & 20,609 & 86,284 & \multicolumn{1}{c|}{0.003} & 0.5 & $32\times64\times20$ \\ \hline
\textbf{{Numerical \& Breast-mimetic phantom}} & $4 \times 16$ & $5 \times 8$ & 53,760 & 291,870 & \multicolumn{1}{c|}{0.002} & 1 & $48\times70\times16$ \\ \hline
\textbf{Mouse(normal)} & $7\times4$& $7\times4$ & 12,288& 63,426 & \multicolumn{1}{c|}{0.0041} & 0.4503 & $32\times32\times12$\\ \hline
\textbf{Mouse (with tumor)} & $7\times7$& $7\times7$ & 12,288& 63,426 & \multicolumn{1}{c|}{0.0045} & 0.3452 & $32\times32\times12$\\ \hline \noalign{\hrule height 0.5pt}
\end{tabular}}
\label{table:dataGen}
\normalsize
\end{table*}

\subsection{{Phantoms and in vivo data}}

To analyze the performance of the proposed approach in controlled real experiments, biomimetic and breast-mimetic phantoms with known inhomogeneity locations was manufactured (see Fig.~\ref{fig:phantom}). The  phantom is made of polypropylene containing a vertically oriented cylindrical cavity that has 20 mm diameter and 15 mm height. The cavity is filled with the acetyl inclusion with different optical properties. 

{For the breast-mimetic phantom, we used a custom-made open-top acrylic chamber (175 mm $\times$ 120 mm $\times$ 40 mm) and three different sized knots (approximately 20 mm, 10 mm, and 5 mm diameter) for the mimicry of a tumor-like vascular structure. The knots were made using thin polymer tubes (I.D 0.40 mm, O.D 0.8 mm diameter) and were filled with the rodent blood that was originated from the abdominal aorta of Sprague-Dawley rat that was under 1 to 2\% isoflurane inhalation anesthesia. The chamber was filled with the completely melted pig lard and the medium was coagulated at room temperature for the imaging scan.}

For in vivo experiment, the mouse colon cancer cell line MC38 was obtained from Scripps Korea Antibody Institute (Chuncheon, Korea) and the cell line was cultivated in Dulbecco's modified Eagle's medium (DMEM, GIBCO, NY, US) supplemented with 10\% Fetal bovine serum (FBS, GIBCO) and 1x Antibiotic-Antimycotic (GIBCO). For the tumor-bearing mice, 5x106 cells were injected subcutaneously into the right flank region of C57BL/6 mice aging 7-9 weeks (Orient Bio, Seongnam, Korea). Animal hairs were removed through trimming and waxing. Anesthesia was applied during the imaging scanning with an intramuscular injection of Zoletil and Rumpun (4:1 ratio) in normal saline solution. Mice were placed inside of the custom-made 80 mm $\times$ 80 mm $\times$ 30 mm open-top acrylic chamber that had a semicircle hole-structure on the one side of the chamber for the relaxed breathing. A gap between the semicircle structure and the head was sealed with the clay. The chamber was filled with the water/milk mixture as 1000:50 ratios. 
All experiments associated with this study were approved by Institutional Animal Care and Use Committees of Asan Medical Center (IACUC no. 2017-12-198).

\subsection{Data preprocessing}


To determine the maximally usable source-detector distance, we measured signal magnitude according to the source-detector distances. 
{We observed that the signals were above the noise floor when the separation distance ($\rho$) between the source and the detector was less than 51 mm ($\rho<51$ mm).} Therefore, instead of using measurements at all source-detector pairs, we only used the pairs having source and detector less than 51 mm apart.
This step not only enhanced the signal-to-noise ratio (SNR) of the data but also largely reduced the number of parameters to be learned in the fully connected layer. 
{For example, in the source-detector configuration of the numerical phantom, the number of input measurement pairs (\#MEAS) reduced from 2560 to 466 (Table \ref{tb:net}).} This decreased the number of parameters to train from 137,625,600 to 25,105,920, which is an order difference. 
To match the scale and bias of the signal amplitude between the simulation and the real data, we multiplied appropriate calibration factor to the real measurement to match the signal envelope from the simulation data. For more detailed information on the measurement data calibration, see Supplementary Material.

\subsection{Neural network training}\label{subsec:dataGen}

To normalize the input data for neural network training, 
we centered the input data cloud on the origin with the maximum width of one by subtracting the mean across every individual data and dividing it by its maximum value. 
To deal with the unbalanced distribution of nonzero values in the 3D label image, 
we weighted the non-zero values by multiplying a constant  scaling factor according to the ratio of the total voxel numbers over the non-zero voxels. 
At the inference stage, the multiplied scaling factor is divided to obtain the true reconstruction value.

\begin{figure}[!hbt]
\centering
\includegraphics[width=0.95\linewidth]{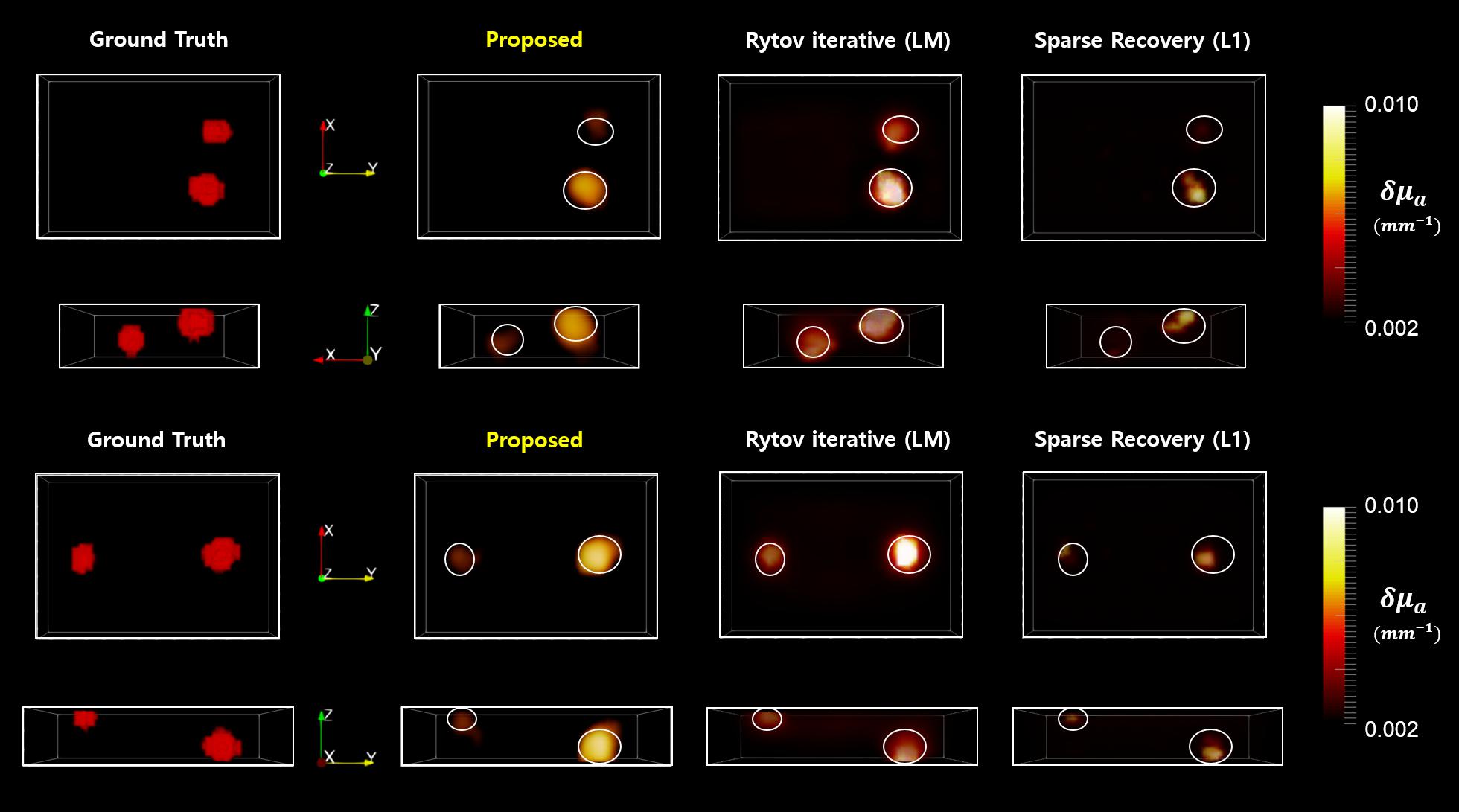}
\vspace*{-0.3cm}
\caption{The reconstruction results from numerical phantom. Here, we used the network trained using the numerical phantom geometry (see Table~\ref{tb:net}). The ground truth images are visualized with binary values to show the location of virtual anomalies clearly. For ease of comparison, the location of the ground truth are denoted by overlapped circles on the reconstructed three dimensional (3D) visualization. }
\label{fig:simul}
\end{figure}

In order to test the robustness of the deep network in real experiments and to obtain a large database in an efficient manner, the training data were generated by solving
diffusion equation using finite element method (FEM) based solver \emph{NIRFAST}
(see, e.g., \cite{dehghani2009near,paulsen1995spatially}). The finite element meshes were constructed according to the specifications of the phantom used in each experiment (see Table~\ref{table:dataGen}). We generated 1500 numbers of data by randomly adding up to three spherical heterogeneities of different sizes (having radii between 2 mm to 13 mm) and optical properties in the homogeneous background. The optical parameters of the heterogeneities were constrained to lie in a biologically relevant range, i.e., two to five times bigger than the background values. 
{For example, in Fig. \ref{fig:simul}, we show two representative reconstructed images among 1500 data that have two inclusions with randomly chosen locations, sizes, and optical parameters.} The source-detector configuration of the data is set to match that of real experimental data displayed in Table~\ref{table:dataGen}.
To make the label in a matrix form, FEM mesh is converted to the matrix of voxels by using triangulation-based nearest neighbor interpolation with an in-built MATLAB $griddata$ function. The number of voxels per each dimension used for each experiment can be found in Table \ref{table:dataGen}.
To train the network for different sizes of phantoms and source-detector configurations, we generated different sets of training data and changed the input and output sizes of the network accordingly. The specifications of the network architecture are provided in Table \ref{tb:net}. Note that 
we intentionally maintained the overall structure of the network same except the specific parameters for consistency and simplicity.

The input of the neural network is the multi-static data matrix of pre-processed measurements. To perform convolution and to match its dimension with the final output of a 3D image, the output of the fully connected layer is set to the size of the discretized dimension for each phantom. All the convolutional layers were preceded by appropriate zero padding to preserve the size. 
As for nonlinearities of our neural network,  we used the hyperbolic tangent function (tanh) as an activation function for the fully connected layer and two convolutional layers (C1 and C2), whereas the last convolutional layer (C3) was combined with rectified linear unit (ReLU) to ensure the positive value for the optical property distribution. 
 {In the network structure for the biomimetic phantom, for example, the first two convolutional layers have 16 filters of $3\times 3 \times 3$ with stride 1 followed by $\tanh$. The last convolutional layer again convolves the filter of $3 \times 3 \times 3$ with stride 1 followed by ReLU (Table \ref{tb:net}).}

We used the mean squared error (MSE) as a loss function and the network was implemented using Keras library \cite{chollet2015keras}. Weights for all the convolutional layers were initialized using Xavier initialization. We divided the generated data into 1000 training and 500 validation data sets. 
For training, we used the batch size of $64$ and Adam optimizer \cite{kingma2014adam} with the default parameters as mentioned in the original paper, i.e., we used learning rate$=0.0001$, $\beta_1 =0.9$, and $\beta_2 =0.999$. Training runs for up to 120 epochs with early stopping if the validation loss has not improved in the last 10 epochs. To prevent overfitting, we added a zero-centered Gaussian noise with standard deviation $\sigma=0.2$ and applied dropout on the fully connected layer with probability $p=0.7$. We used a GTX 1080 graphic processor and i7-6700 CPU (3.40 GHz). The network took about 380 seconds for training. 
Since our network only used the generated simulation data for training, there are potential  that the network could  be  suffered from the noise not observed from the synthetic data. However, by careful measurement data calibration and the parameter tuning for the dropout probability and the standard deviation $\sigma$ of the Gaussian noise, we could achieve the current network architecture which performs well in various experimental situations. We have not imposed any other augmentation such as shifting and tilting since our input data are not in the image domain but in the measurement domain which is unreasonable to apply such techniques.
Every 3D visualization of the results is done by using ParaView \cite{ahrens2005paraview}.

The simulation data and code will be available on authors' homepage (http://bispl.weebly.com) upon publication. 

\subsection{Baseline algorithm for comparison}

As the baseline methods, we employed two widely used algorithms that are implemented in the state of the art public software packages of DOT field (time-resolved optical absorption and scattering tomography (TOAST) \cite{schweiger2014toast++} and Near Infrared Fluorescence and Spectral Tomography (NIRFAST) \cite{dehghani2009near}). One is a distorted Rytov iterative method. This algorithm employs the modified Levenberg Marquardt (LM) algorithm. The other is based on a penalized least squares recovery method with various sparsity inducing penalty that employs the homotopy-like cooling approach with a help of majorization minimization (MM) framework \cite{prakash2014sparse}.  
In these algorithm, at each iterative step we re-calculate  Green's function along with update of unknown parameter values.
We set the convergence criterion if the reconstructed optical parameter at the current iteration has not improved in the last two iterations. Unless an initial guess is bad, the algorithms generally converged in six to ten iterations and each iteration took approximately 40 seconds, which makes total reconstruction time about {7 minutes}.

{The regularization parameter of the LM method and the penalized least squares algorithm is chosen as $\lambda = constant\times \max(\mathrm{diag}(J^T J))$ where $J$ is the system matrix or Jacobian matrix and  $\mathrm{diag}$ refers to the vector composed of diagonal elements. The value of the constant used for reconstruction are found by trial and error. 
}
For the penalized least squares algorithms, we compared the performance with two $\ell_p$ norm penalties of $p=1,2$, where
$p=1$ corresponds to the sparsity inducing penalty.  

\begin{figure*}[!hbt]
\centering
\includegraphics[width=0.75\linewidth]{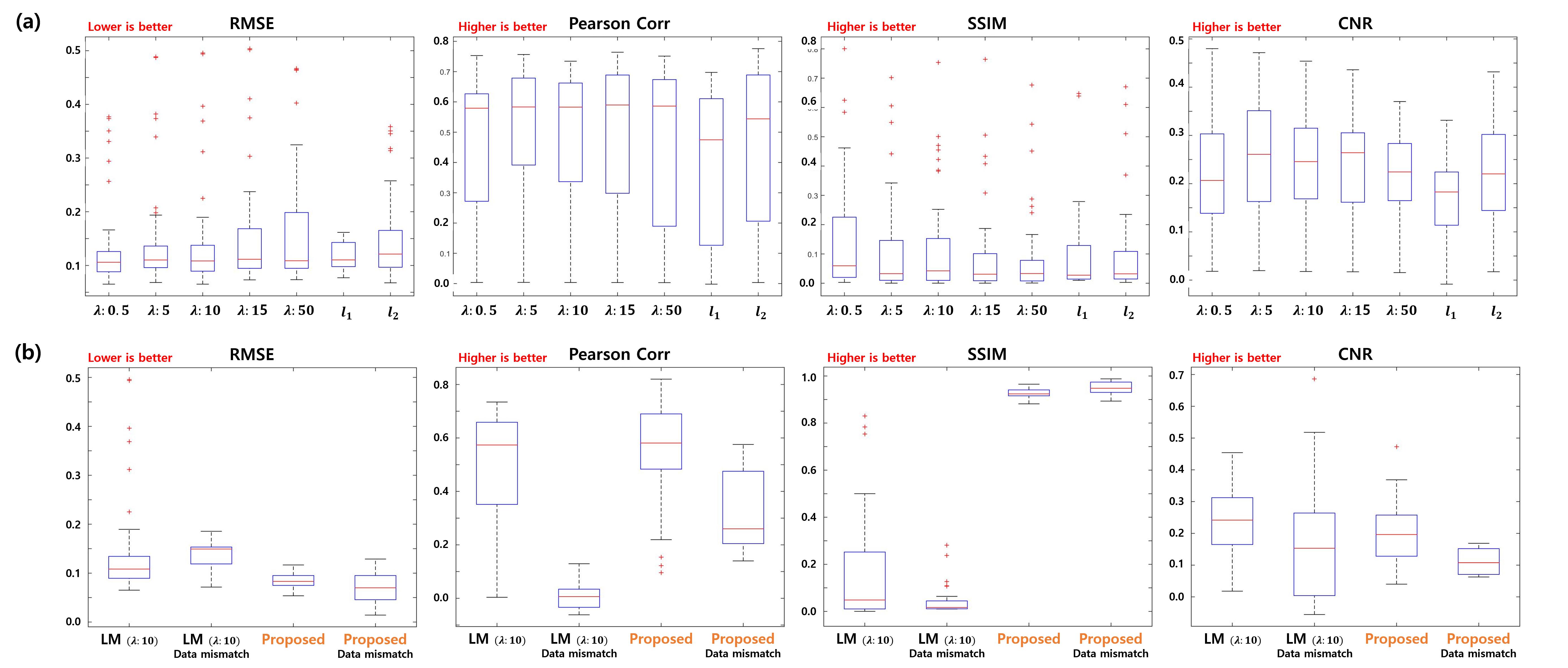}
\caption{(a) Quantitative comparisons of   the various baseline  algorithms different hyperparameters. (b) Comparison of the baseline algorithm and
 the  proposed method under 
 boundary condition mismatch. Root mean square error (RMSE), Pearson's correlation, structural similarity index (SSIM) and contrast-to-noise ratio (CNR) are evaluated on randomly chosen 38 images from the test reconstruction of the numerical phantoms. }
\label{fig:quant}
\end{figure*}

\begin{figure*}[!hbt]
\centering
\includegraphics[width=0.8\linewidth]{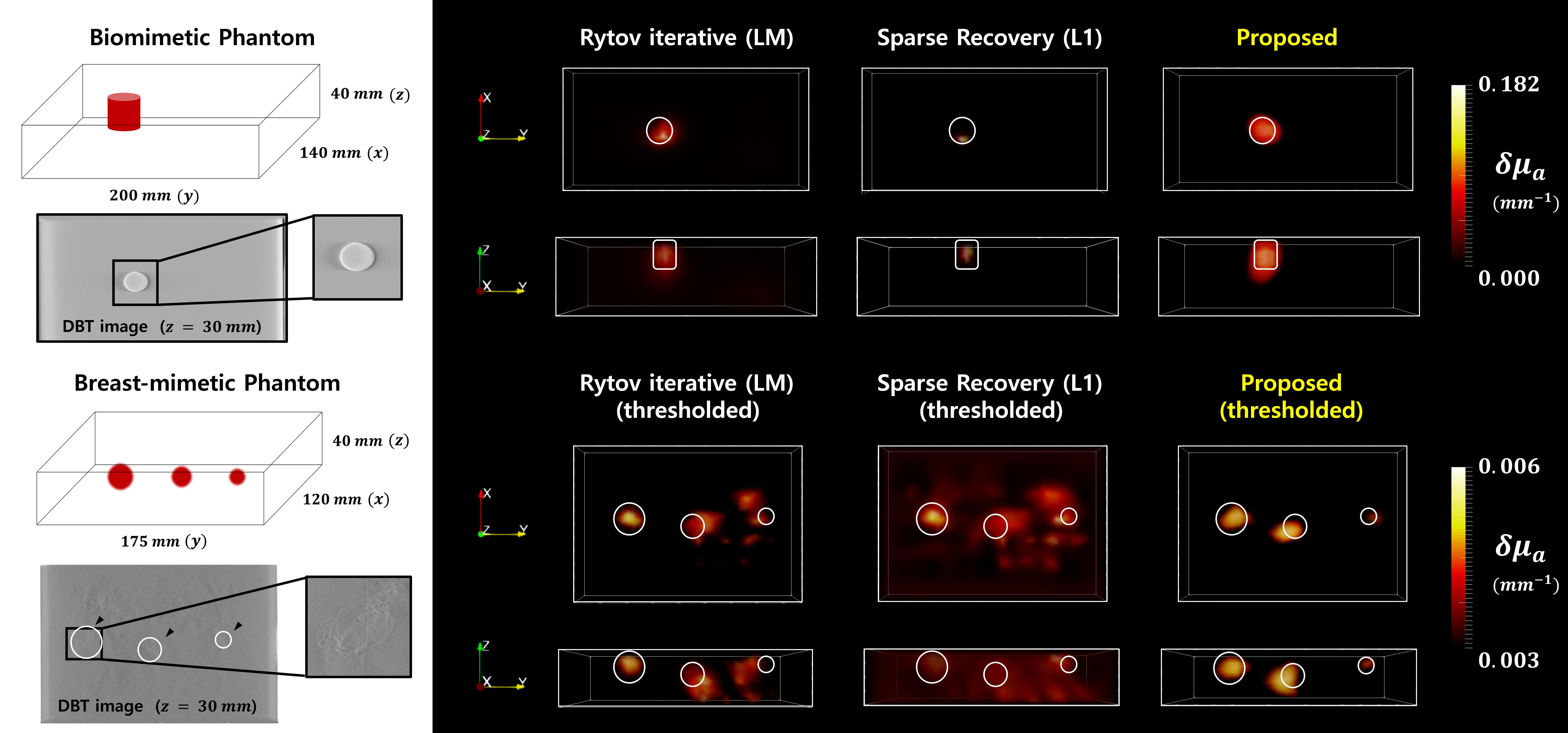}
\caption{{The reconstructed images of the biomimetic and breast-mimetic phantoms. 
The contrast can be seen by the DBT images. For the biomimetic phantom experiments, to provide more clear results of the conventional methods, images are thresholded to remove the small intensities. Every image is the result of summed projection of the intensity values along the viewpoint. The ground truth location of the single inclusion model and the \textit{expected} ground truth location of the breast-mimetic model are denoted by white boundaries within the reconstructed images.
}
}
\label{fig:phantom}
\end{figure*}

\section{Results}
\label{sec:Results}

To validate the algorithms in a quantitative manner, we evaluated four metrics such as the root mean square error (RMSE), Pearson correlation, structural similarity index (SSIM) and contrast-to-noise ratio (CNR) on the reconstructed results of randomly chosen 38 data configurations with different anomaly sizes and locations, with distinct z-locations. For more details on CNR formula, please refer to our Supplementary Material. 

First, we conducted simulation study using  numerical phantoms  shown in the leftmost column of Fig.~\ref{fig:simul}.
In Fig.~\ref{fig:quant}, we also report the quantitative performance  comparison results using box plot to show the overall statistics. On each box, the central mark is the median, the edges of the box are the 25th and 75th percentiles, the whiskers extend to the most extreme data points not considered outliers, and outliers are plotted individually.
As shown in Fig.~\ref{fig:quant}(a), we could not find a significant difference in the performance of the conventional methods over  varying hyper-parameters within the specific range. 
{Based on the quantitative results, in this paper, we mainly used $\lambda=10$  for both LM method and the sparse recovery algorithm with $\ell_1$ norm throughout the paper.
}

Next, to analyze the robustness of the proposed approach, we evaluated the method on the mismatching boundary condition scenario (Fig.~\ref{fig:quant}(b)). More specifically, the {training data of our network is generated based on the condition considering the refractive index  mismatch at the boundary while} the 
test data is generated using the boundary condition that are not matched to the training data {so that the initial assumption breaks down at the test time.} 
To avoid the inverse crime,
 we used more dense FEM and reconstruction mesh for generating forward data than the reconstruction mesh.
As shown in Fig.~\ref{fig:quant}(b), the performance of the proposed method is better than the conventional method. More specifically, in the RMSE plot, the result of the proposed method did not vary a lot  while the RMSE of the conventional method increased when the boundary condition differs from the initial assumption. In particular,
in  Pearson correlation and SSIM values, the proposed method significantly outperformed the conventional methods. 
Some representative reconstruction results by various methods are shown in Fig.~\ref{fig:simul}, which clearly confirm the effectiveness of the proposed method.

{To analyze the performance of the proposed approach under more realistic and controlled environments, biomimetic and breast-mimetic phantoms with known inhomogeneity locations were examined (see Fig.~\ref{fig:phantom}). We obtained the measurement data using our multi-channel system. The reconstructed 3D images from the conventional methods (LM and sparse recovery with $l_1$ penalty) and our proposed network are compared.}

{
The biomimetic phantom provides a simple tumor model with a vertically oriented cylindrical inclusion that has different optical properties to the background.
Here, both the conventional LM method ($\lambda=10$) and our network reconstructed the location of optical anomalies, among which the proposed method has more accurate reconstruction (Fig.~\ref{fig:phantom}). The sparse recovery based algorithm ($\lambda=10$) were able to find the center location but it suffered from underestimation of the support due to the sparsity constraint. 
Because the phantom has high contrast between the inclusion and its background, the contrast can be clearly seen in the DBT image as well. }

\add{Next, we examine a more complicated tumor model using the breast-mimetic phantom, which is more realistic than the biomimetic phantom. Specifically,   because the phantom includes inhomogeneous backgrounds (coagulated pig lards) and the inclusions are made of thin polymer tubes filled with the rodent blood, it provides additional technical challenges  beyond  the commonly used assumptions such as a homogeneous background and known boundary conditions.  Moreover, due to its ingredients, the contrasts cannot be clearly seen by the DBT image  (Fig.~\ref{fig:phantom}). 
}

{
Because of these complicated optical characteristics, for the breast-mimetic phantom experiment, the conventional methods suffer from the strong background noises that appear near the sensor plane at the bottom. 
Even after applying post-processing to threshold out the small intensity values, the conventional methods show artifacts in the recovered image, which is more prominent in the sparse recovery reconstruction. 
 Unlike the conventional methods, our proposed method recovers the locations of inclusions accurately. Here, just for a fair comparison, we applied the thresholding with the same range on the recovered image using our method, although our method could recover the inclusions only, even without the post-processing. Moreover,  our method can accurately recover their relative sizes of the inclusions. Although this can be only shown in qualitative way due to the lack of the ground truth, the results show the favorable characteristics of our methods clearly. }

\begin{figure*}[!hbt]
\centering
\includegraphics[width=0.7\linewidth]{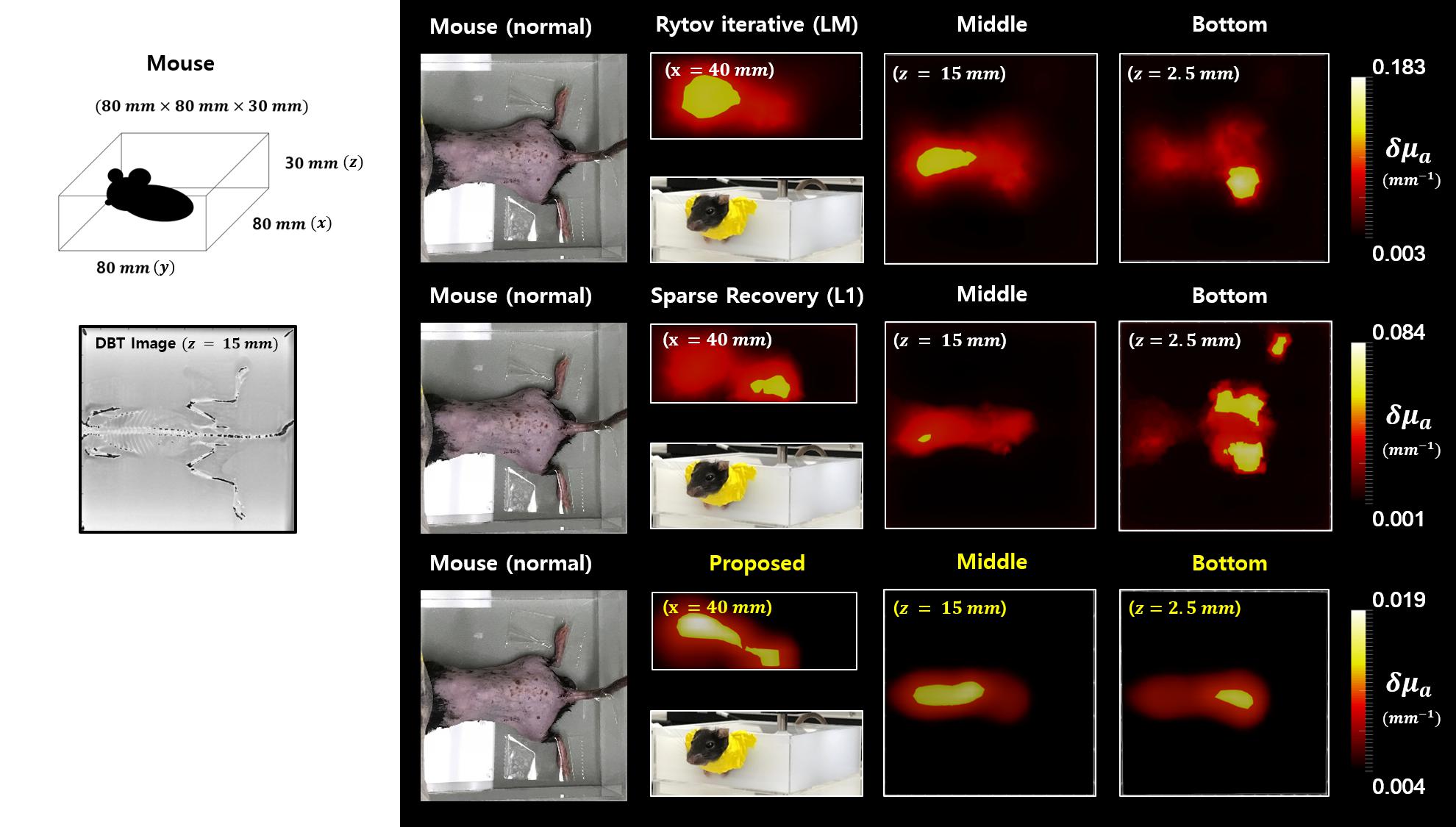}
\caption{The reconstructed images of \emph{in vivo} animal experiments. Mouse without any tumor before water/milk mixture was filled is shown. 
In order to get the scattered photon density measurements, we recorded the data with and without the mouse placed in the tank, which is filled with the water/milk mixture of $1000:50$ ratio. 
Both the conventional and the proposed methods recovered high $\delta\mu$ values at the chest area of the mouse. {However, the LM method   finds a big chunk of high $\delta\mu$ values around the left thigh of the mouse, and the sparse recovery method finds those in both thighs with an additional spurious artifact outside of the mouse body, which is unlikely with the normal mouse. In contrast, our proposed network shows a high $\delta\mu$ along the spine of the mouse where the artery and organs are located.} }
\label{fig:invivo-homo}
\end{figure*}

\begin{figure*}[!hbt]
\centering
\includegraphics[width=0.7\linewidth]{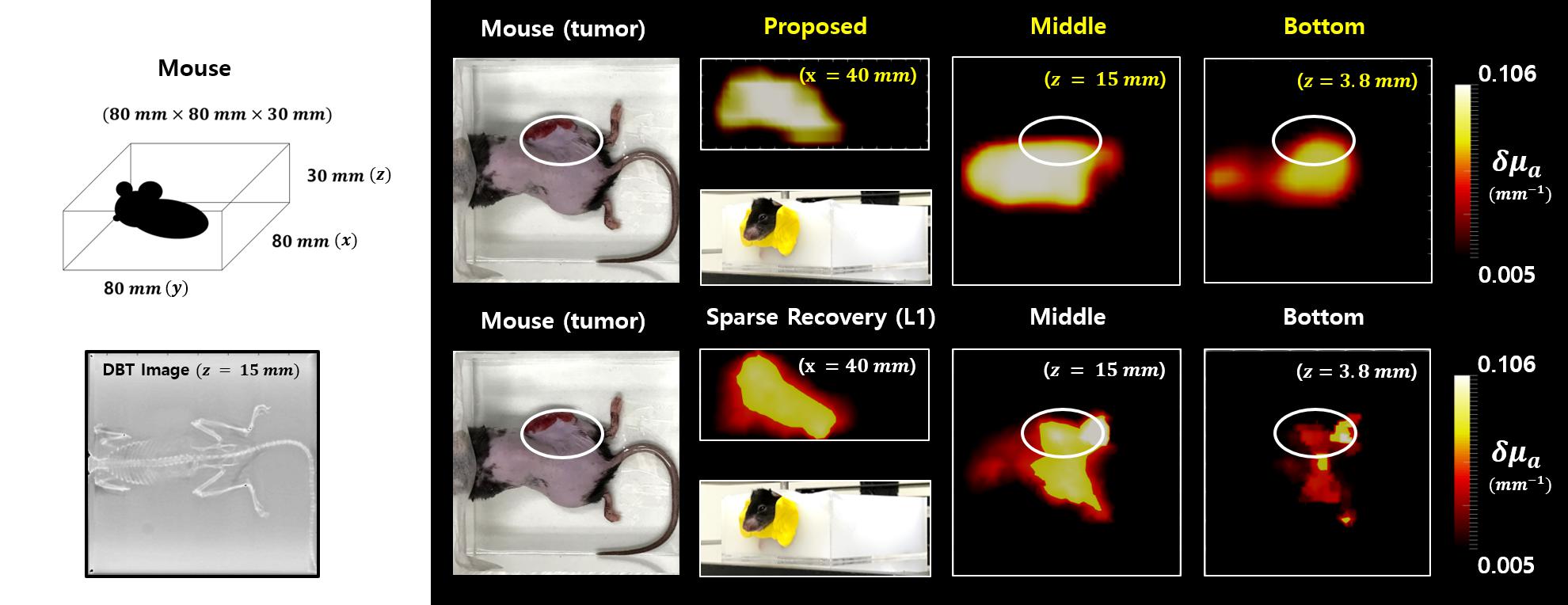}
\caption{The reconstructed images of \emph{in vivo} animal experiments. Mouse with tumor on the right thigh before water/milk mixture was filled is shown. Comparison between the mouse with and without tumor 3D visualizations are displayed. In addition, we showed a slice in $x$ and $z$ directions for a clear visualization. {When compared with a normal mouse experiment (Fig.~\ref{fig:invivo-homo}), our network finds a high $\delta\mu$ values around the right thigh, where the tumor is located.}
}
\label{fig:invivo-inhomo}
\end{figure*}

Finally, we performed \emph{in-vivo} animal experiments using a mouse (Fig.~\ref{fig:invivo-homo} and Fig.~\ref{fig:invivo-inhomo}). In order to get the scattered photon density measurements, we recorded the data with and without the mouse placed in the chamber, which is filled with the water/milk mixture of $1000:50$ ratio. 
Optical scattering data from animal experiments were collected using the single-channel system. Fig.~\ref{fig:invivo-homo} and Fig.~\ref{fig:invivo-inhomo} show the reconstructed images of the mouse with and without tumor. The conventional algorithms and the proposed methods recovered high $\delta\mu$ values at the chest area of the mouse. {However, the LM and the sparse recovery reconstruction finds a big chunk of high $\delta\mu$ values around the thighs of the mouse, which is unlikely with the normal mouse (Fig.~\ref{fig:invivo-homo}).} In contrast, our proposed network shows a high $\delta\mu$ along the spine of the mouse where the artery and organs are located \cite{cheong1990review}. Furthermore, in the mouse with tumor case, our network finds a high $\delta\mu$ values around the right thigh, where the tumor is located (Fig.~\ref{fig:invivo-inhomo}) \cite{fang2011combined}.
\add{On the other hand,  the sparse recovery method outputs highest intensity values around its feet and bladder.} The lateral view of our reconstructed images also matches with the actual position of the mouse, whose head and body are held a little above the bottom plane due to the experiment setup.

\section{Discussion}

{Compared to the results of the conventional LM method and the sparse recovery algorithm}, our network showed  robust performance over the various examples. 
While the conventional reconstruction algorithms often imposed high intensities on spurious locations (Fig.~\ref{fig:phantom} bottom row) and Fig.~\ref{fig:invivo-homo} top and middle row), our network found accurate positions with high values only at the locations where inclusions are likely to exist. This is because unlike the conventional method that requires a parameter tuning for every individual case, our network can infer from the measured data without additional pre- and post-processing techniques. 

Note that our network had not seen any real data during the training nor the validation process. However, it successfully finds the inversion by learning only from the simulation data. Furthermore, even though we trained the network with examples having sparse supports, our network successfully finds both sparse (phantom, Fig.~\ref{fig:phantom}) and extended targets (mouse, Fig.~\ref{fig:invivo-homo} and Fig.~\ref{fig:invivo-inhomo}) without any help of regularizers. {These results evidence that our network can learn a general inverse function instead of learning a trivial mapping (memorization). Both of these results are very surprising in the perspective of learning a function using a parametric model with large size, such as the neural networks, which are usually known to easily overfit (memorize) to the training data.} 

{We argue that this surprising generalizability comes from 1) the regularization techniques we employed for training and 2) the specific design of our network.} 
Recall that we added the stochastic noise to the data when we train the network. In addition, ``dropout'' adds another randomness to the network, which randomly drops the nodes of the network while training. Both of Gaussian noise addition and dropout layers make the network to generalize better on the unseen data or conditions, which adds another practical advantage of our method. This is one of the reasons that enabled our proposed network to perform well on the real data even though it had not seen any real data during the training phase. 

{Secondly, our network is designed to directly learn from the measurement data, and does not start from the reconstructed images using the conventional methods that inevitably requires a prior conditions that might bias the search space. Therefore, it can generalize better on the unseen conditions, such as the boundary condition mismatch (Fig. \ref{fig:quant}).}

\begin{figure}[!hbt]
\centering
\includegraphics[width=1\linewidth]{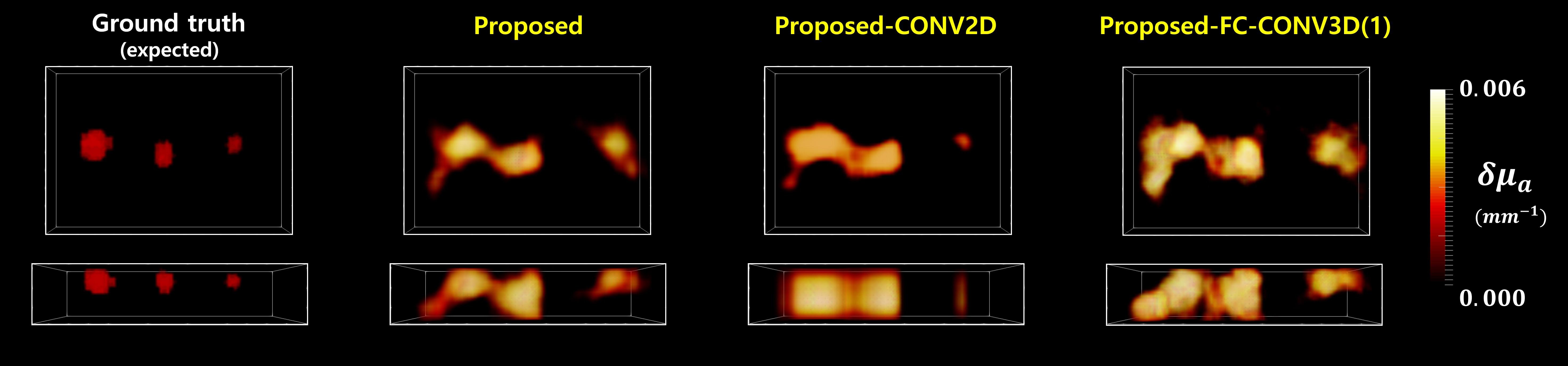}
\caption{{Network ablation study results. The reconstructed images of the breast-mimetic phantom (Fig. \ref{fig:phantom} bottom row) using the networks with 2D- and 3D-convolution are compared (the 2nd and 3rd columns).} To show the necessity of the convolutional layers, the image reconstructed using the network with a single 3D-convolution layer of a single filter is shown (last column). Meanwhile, the network with only a fully connected layer failed to train (results not shown). Every image is the result of summed projection of the intensity values along the viewpoint.}
\label{fig:ablation1}
\end{figure}

{To further show that the main performance of our method does not come from  heuristically fine-tuned hyperparameters (number of the layers, channels, etc.) but from its sequential architecture of the fully connected layer followed by the convolutional encoder-decoder layers, we performed ablation studies by changing or removing the components of the proposed architecture.} Since our output $f$ is a 3D distribution, the network needs to find a set of 3D filters $\alpha(\Psi)$ and $\beta(\tilde{\Psi})$. 
We observed that the network with 3D-convolution showed better $z-$axis resolution compared to the one using 2D convolution (Fig.~\ref{fig:ablation1}), which is consistent with our theoretical prediction.
One may suspect that the performance of the network has originated solely from the first layer since over 98\% of the trainable parameters are from the fully connected layer. To address this concern, we tested the network with and without convolutional layers after the fully connected layer. We observed that the performance of our network deteriorated severely and it failed to train without the consecutive convolution layers (the results not shown). At least a single convolution layer with a single filter was needed to recover the accurate location of the optical anomalies (Fig.~\ref{fig:ablation1}). 
{Indeed, the inverted output of the fully connected layer shows a very noisy reconstruction, which is then refined by the encoding and decoding procedure of consecutive convolutional layers (see Supplementary Material). 
However, the paired encoder-decoder filters in the proposed network are better than just using a single convolution layer. 
} 

Note that a similar fully connected layer $\mathcal{T}$ in \eqref{eq:enc-dec}  at the first stage was investigated in \cite{zhu2017image}.
However, for the case of  magnetic resonance imaging (MRI) or X-ray computed tomography (CT) problems, there exists  well-defined inversion layers  using Fourier transform or filtered backprojection, so analytic reconstruction is recommended  rather than  using the fully connected layers \cite{kang2017deep,jin2017deep,antholzer2017deep,yoon2018efficient}, since the number of unknown parameters for the fully connected layers are too big.
On the other hand, for inverse scattering problems such as DOT,  the analytic reconstruction is not well-defined, so 
the fully connected inversion layer can be also
estimated from the data as demonstrated in this paper.
\add{Of course, for the case of well-defined numerical simulation, one could have a good initial reconstruction from the LM type reconstruction, so standard convolutional neural network architecture could be trained as a denoiser for the LM reconstruction.  However, for the DOT problem from real data, the initial guess using LM method is seriously dependent on the boundary conditions and regularization parameters due to the mismatch between the real and numerical Green’s kernel. Thus, learning from such unstable initial guess is not robust. This is why we are interested in learning the inverse as well using a fully connected layer. 
}
 
%
%

There are several limitations  in the current study.  
Although the theory provides a useful global perspective to design the network architecture, it does not still answer the DOT specific questions, such as specific type of nonlinearities, filter sizes, and etc.  Still, providing the global perspective to network architecture is also important since it reduces the search space of the neural network to specific details that can be readily obtained by trial and error.  
In terms of hardware design, the present hardware system  also has some  limitations.
First, the detection relies on few fibers. CCD based-detection allows better sampling (spatial resolution) and is non-contact, but fiber measurements also suffer from coupling problems.
Second, measurements are performed filling the imaging tank with liquid. Adding liquid has several side effects such as increases of the scattering 
as well as the reduction of detected photon, resulting in signal-to-noise ration loss.  
Accordingly, the fiber-based measurement have intrinsic limitations, so the proposed reconstruction algorithm, even though it is better than the other reconstruction
method,   may also suffer from such hardware limitation.
Fortunately, the proposed method is quite general, so it  can be  easily modified for different hardware systems.

Note that the main challenges that limit wide uses of DOT are: 1) to recover large absorption changes, 
2) to be robust to light propagation model deviations, and 3) to dispose of background measurements (absolute vs relative reconstruction).
Recent work for optical diffraction tomography \cite{sun2018efficient} provided convincing results that deep neural network can address multiple scattering from large perturbation.
In addition, the direct inversion based on the  Lippman-Schwinger formulation is shown to address multiple scattering and provide super-resolution
thanks to the internal resonance mode enhancing \cite{lim2017beyond}.
Thus, 
 the proposed approach can potentially solve  the first two issues.
Unfortunately,  our method does not still address the third problem, since our the Lippman-Schwinger integral
formulation requires the flux measurement from homogeneous background.
In many clinical applications such as breast cancer imaging, such additional measurement from homogeneous background is difficult to obtain.
Thefore, the extension of the proposed learning approach for such practical environment would be an interesting research direction,
 but is  beyond the scope of current work. 

\section{Conclusion}
\label{sec:Conclusion}
In this paper, we proposed a deep learning approach to solve the inverse scattering problem of diffuse optical tomography (DOT). Unlike the conventional deep learning approach, which tries to denoise or remove the artifacts from image to image using a black-box approach for the neural network, our network was designed based on Lippman-Schwinger equation to learn the complicated non-linear physics of the inverse scattering problem. Even though our network was only trained using the  numerical data, we showed that the learned network
provides improved reconstruction results over the existing approaches in both simulation and  real data experiments
and accurately reconstructs the anomalies without iterative procedure or linear approximation. 
By using our deep learning framework, the non-linear inverse problem of DOT can be solved in end-to-end fashion and new data can be efficiently processed in a few hundreds of milliseconds, so it would be useful for dynamic imaging applications. 
\bibliographystyle{IEEEtran}
\bibliography{pnas-dot,convframelets}
\appendices

\section{Measurement Data Calibration}

\begin{figure}
\centering
\includegraphics[width=1\linewidth]{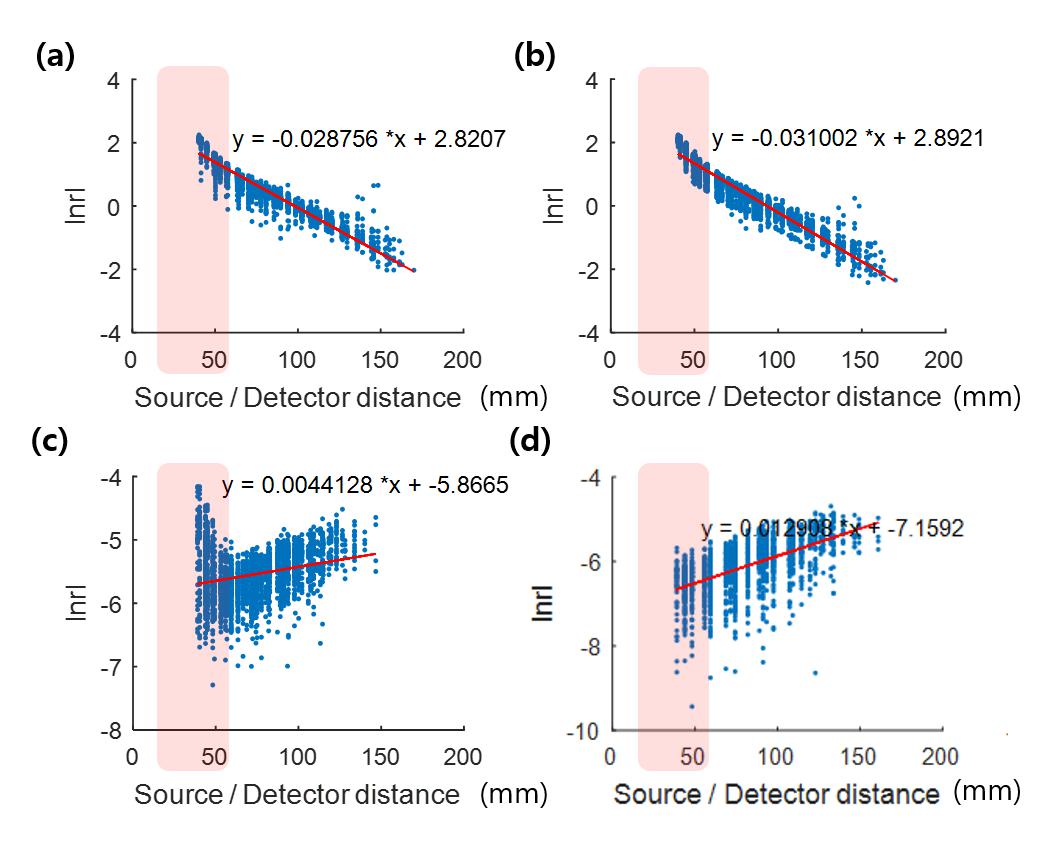}
\caption{Plot of the signal amplitude of all possible source-detector combinations versus the corresponding source-detector distances measured in different experiments. (a) Polypropylene phantom with one inclusion. (b) Polypropylene phantom with three inclusions. (c) Animal (normal). (d) Animal (tumor).}
\label{fig:src-det-amp}
\end{figure}

To determine the maximally usable source-detector distance and investigate the variation of the real measurement data, 
we measured every phantom and calibrated signal magnitude versus source-detector distances. The resulting plot can be seen in Figure~\ref{fig:src-det-amp}. From this figure, for every measurement, we observe that the separations of less than $\sim 51~mm$ result in detectable signals above the noise floor, whereas noise dominates in animal experiment measurements with source-detector distances of more the $51~mm$. For animal (tumor) experiment, even for source-detector distances $<51 mm$, the signal level does not seem to decrease as source-detector distance increases. Still, if zoomed, the signal level decreases and goes up (V-shaped) in the range of source-detector distances $<51~mm$. 
\textcolor{black}{Since our goal is to use the same trained neural network for various experiments, for consistency we performed the same preprocessing of the animal data as we did for  the other experiments.  }

The image reconstruction process starts from estimating the bulk optical properties of the medium. If the bulk optical properties are incorrectly estimated, this inaccurate starting point may lead to slow convergence or converge to incorrect readings. The bulk optical properties are calculated by assuming the heterogeneous medium as uniform bulk medium. The uniform bulk optical properties are found by fitting the experimental data to the model based data (diffusion equation in this case) using iterative Newton Raphson scheme  as suggested in \cite{pogue2000calibration}.

\begin{figure}[!hbt]
\centering
\includegraphics[width=1\linewidth]{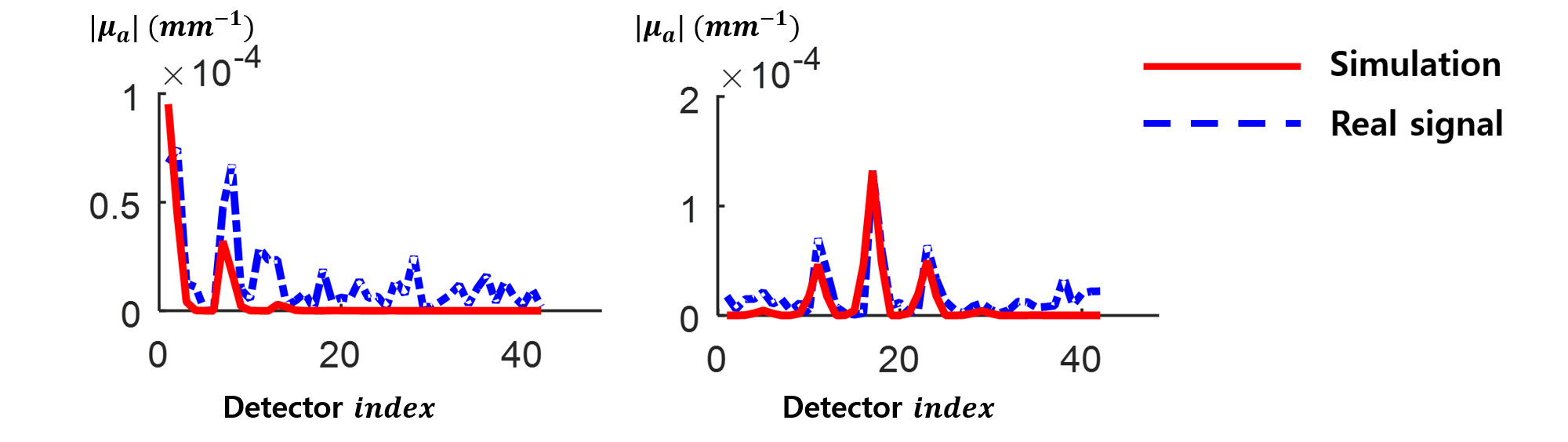}
\caption{An example of before (left panel) and after (right panel) pre-processing for matching the training data profile to the real signal envelop. $x$-axis denotes the detector indices with respect to a fixed source location and $y$-axis denotes the absolute $\mu_a $ value $(mm^{-1})$.}
\label{fig:data_fit}
\end{figure}

To match the scale and bias of the signal amplitude between simulation and real data, we divided and added an appropriate constant value to the simulation data to match the maximum peaks.
Examples of the matched measurement and numerical data  after the calibration are illustrated in  Figure~\ref{fig:data_fit}.
Note that the measurement still contain lots of noisy, so the neural network should be robustly trained using additive noise and dropouts.


\section{Hyper-parameter Selection} 	
\label{sec:results}
Since it is difficult to perform an exhaustive architecture search to  the optimal network structure for each experimental
condition, 
for consistency and simplicity we intentionally maintained the overall structure of the network same for all the data set except the specific parameters. Specifically, we used the same architecture (fixed number of layers) because we wanted to show the unified architecture that generally works over the different data (poly phantom, biomimetic phantom, and animal real data) by just changing the number of channels.

In the following, we illustrate experimental results for hyper-parameter selection that was performed for the numerical phantom dataset (Fig. 3). 
%
%
%

\subsection{Number of intermediate convolutional layers for denoising}
As it can be seen in Table \ref{tb:addlayer}, an additional denoising layer $H$ of 64 channels (Model 1) to the original numerical model (baseline) did not provide any significant improvement in the model performance (RSME, Pearson's Correlation, SSIM and CNR) while it adds another 110,592 parameters to the model risking the overfitting. (Please refer to subsection \ref{subsec:cnr} for detailed calculation procedure of CNR). 
A lighter architecture without denoising layer (Model 2) can also work fine for this simple problem. 
Therefore, we chose a single denoising layer as a compromise for general problems,  since this was the maximum number of layers we could mount the model on a single GPU. 

Note that the proposed neural network is memory intensive due to the fully connected layer. 
For example, among the total 25,219,968 number of parameters of the network for the numerical phantom experiment, the fully connected layer occupies 25,105,920 (99.55\%) number of the parameters. 
This limited the maximum number of convolution layers due to the physical limits of the GPU memory size. 
\begin{table}[!hbt]
\centering
\caption{The variations of the model performance over different number of denoising layers (mean $\pm$ standard deviation). 
Model 1 and Model 2 are constructed by adding and subtracting one convolutional layer with 64 channels
from the baseline, respectively.}
\label{tb:addlayer}
\resizebox{0.41\paperwidth}{!}{
\begin{tabular}{|c|c|c|c|}
\hline
                  & \textbf{Baseline} & \textbf{Model 1} & \textbf{Model 2} \\ \hline
\textbf {\# denoising layers} &  1                   & 2        & 0        \\ 
\textbf{RMSE}     & 0.0702 $\pm$ 0.0291  & 0.0709 $\pm$ 0.0296 & 0.0700 $\pm$ 0.0289\\ 
\textbf{Pearson's Corr.}     & 0.5368 $\pm$ 0.1910 & 0.5359 $\pm$ 0.1676 & 0.5471 $\pm$ 0.1910 \\ 
\textbf{SSIM}     & 0.9425 $\pm$ 0.0263& 0.9421 $\pm$ 0.0266 & 0.9419  $\pm$ 0.0269 \\ 
\textbf{CNR}     & 0.1842 $\pm$ 0.0859  &0.1804 $\pm$ 0.0793 & 0.1899 $\pm$ 0.0892 \\ \hline
\end{tabular}
}
\end{table}
%
\subsection{Choice of the number of filter channel $r$}
The rank $r$ of the Hankel matrix corresponds to the number of the convolution channels of the network. 
In order to see the dependency on $r$,
 we additionally trained the network with varying number of channels. By reducing or increasing the number of the channels, the variations of the network performance were marginal or negligible (Table \ref{tb:addfilter}). 
This relatively insensitiveness to the rank $r$ is originated from simplicity of abnormality used in this experiment. Because all the inclusions are numerically generated, smooth and spherical-shaped objects, they can be easily fitted by using low rank filters.

\subsection{MATLAB code for CNR calculation}
\label{subsec:cnr}
For clarity, we provide a matlab code for calculating contrast-to-noise ratio (CNR). 
\begin{table}[!h]
\begin{tabular}{|l|}
\hline
\textbf{MATLAB code for CNR calculation}\\ \hline
\begin{tabular}[c]{@{}l@{}}\textcolor{ao(english)}{\% clabel is binary matrix of the ground truth} \\ \textcolor{ao(english)}{\% cestimate is the estimated matrix of the ground truth} \\
ind\_roi = find(clabel(:)$\sim$=0);\\ ind\_back = find(clabel(:)==0); \\ a\_roi = size(ind\_roi,1)/size(clabel,1); \\ a\_back = size(ind\_back,1)/size(clabel,1);\\ mean\_roi = mean(cestimate (ind\_roi));\\ var\_roi = var(cestimate (ind\_roi));\\ mean\_back = mean(cestimate (ind\_back));\\ var\_back = var(cestimate(ind\_back));\\ CNR = (mean\_roi-mean\_back) / \\ \qquad\quad (sqrt(a\_roi*var\_roi+a\_back*var\_back));\end{tabular} \\ \hline
\end{tabular}
\end{table}
\begin{table*}[!hbt]
\centering
\caption{The variations of the model performance over different number of channels (mean $\pm$ standard deviation). Best performance is marked by \textbf{bold-face}.}
\label{tb:addfilter}
\resizebox{0.53\paperwidth}{!}{
\begin{tabular}{|c|c|c|c|c|c|}
\hline
                  & \textbf{Baseline} & \textbf{Model 1} & \textbf{Model 2} & \textbf{Model 3} & \textbf{Model 4}\\ \hline
\textbf{\# filters ($r$) } & $64$ & $80$ & $32$ & $16$ & $1$        \\ 
\textbf{RMSE}     & \textbf{0.0702 $\pm$ 0.0291} & 0.0703 $\pm$ 0.0293 & 0.0703 $\pm$ 0.0294 & 0.0703 $\pm$ 0.0292 & 0.0708 $\pm$ 0.0296\\ 
\textbf{Pearson's Corr.}     & 0.5368 $\pm$ 0.1910 & 0.5645 $\pm$ 0.1673 & 0.5267 $\pm$ 0.2113& \textbf{0.5647 $\pm$ 0.1499} & 0.5431 $\pm$ 0.1676 \\ 
\textbf{SSIM}     & 0.9425 $\pm$ 0.0263 & 0.9423 $\pm$ 0.0267 & 0.9427  $\pm$ 0.0264 &\textbf{ 0.9427 $\pm$ 0.0263} & 0.9418 $\pm$ 0.0271\\ 
\textbf{CNR}     & 0.1842 $\pm$ 0.0859 &\textbf{0.1985 $\pm$ 0.0951} & 0.1842 $\pm$ 0.1028 & 0.1942 $\pm$ 0.0800 & 0.1847 $\pm$ 0.0820 \\ \hline
\end{tabular}
}
\end{table*}

\begin{table*}[!hbt]
\centering
\caption{The dependency on SNR of the model performance (mean $\pm$ standard deviation). Best performance is marked by \textbf{bold-face}.}
\label{tb:snrdep}
\resizebox{0.75\paperwidth}{!}{
\begin{tabular}{|c|c|c|c|c|c|c|}
\hline
                  & \textbf{w/o noise} & \textbf{SNR 0 dB} & \textbf{SNR 1 dB} & \textbf{SNR 5 dB} & \textbf{SNR 10 dB} & \textbf{SNR 20 dB}\\ \hline
\textbf{RSME}     & \textbf{0.0702 $\pm$ 0.0291}  & 0.0707 $\pm$ 0.0296 & 0.0710 $\pm$ 0.0298 & 0.0703 $\pm$ 0.0293 & 0.0704 $\pm$ 0.0293 & 0.0703 $\pm$ 0.0293 \\
\textbf{Pearson's Corr.}     &\textbf{0.5368 $\pm$ 0.1910 }& 0.5101 $\pm$ 0.2163 & 0.5309 $\pm$ 0.1735 & 0.5239 $\pm$ 0.2122 & 0.5227 $\pm$ 0.2138 & 0.5244 $\pm$ 0.2127 \\
\textbf{SSIM}     & \textbf{0.9425 $\pm$ 0.0263}& 0.9422 $\pm$ 0.0265 & 0.9422 $\pm$ 0.0267 & 0.9423 $\pm$ 0.0267 & 0.9422 $\pm$ 0.0267 & 0.9422 $\pm$ 0.0267 \\
\textbf{CNR}     & \textbf{0.1842 $\pm$ 0.0859} & 0.1746 $\pm$ 0.0947 & 0.1792 $\pm$ 0.0811 & 0.1807 $\pm$ 0.0939 & 0.1803 $\pm$ 0.0944 & 0.1810 $\pm$ 0.0940 \\ \hline
\end{tabular}
}
\end{table*}

\subsection{Dependency on SNR}
To show that our model is robust over different signal-to-noise ratio (SNR), we prepared the noisy measurement data by adding the white Gaussian noise with varying SNR in decibel (dB) scale (Table \ref{tb:snrdep}). The noisy measurements were used at the test stage, even though our neural network was only trained on the clean measurement without noise. The variations of the model performance were marginal across different SNR values, which clearly confirms the robustness of the algorithm.

\subsection{Role of the fully connected layer}
\begin{figure}[!hbt]
\centerline{\includegraphics[width=8cm]{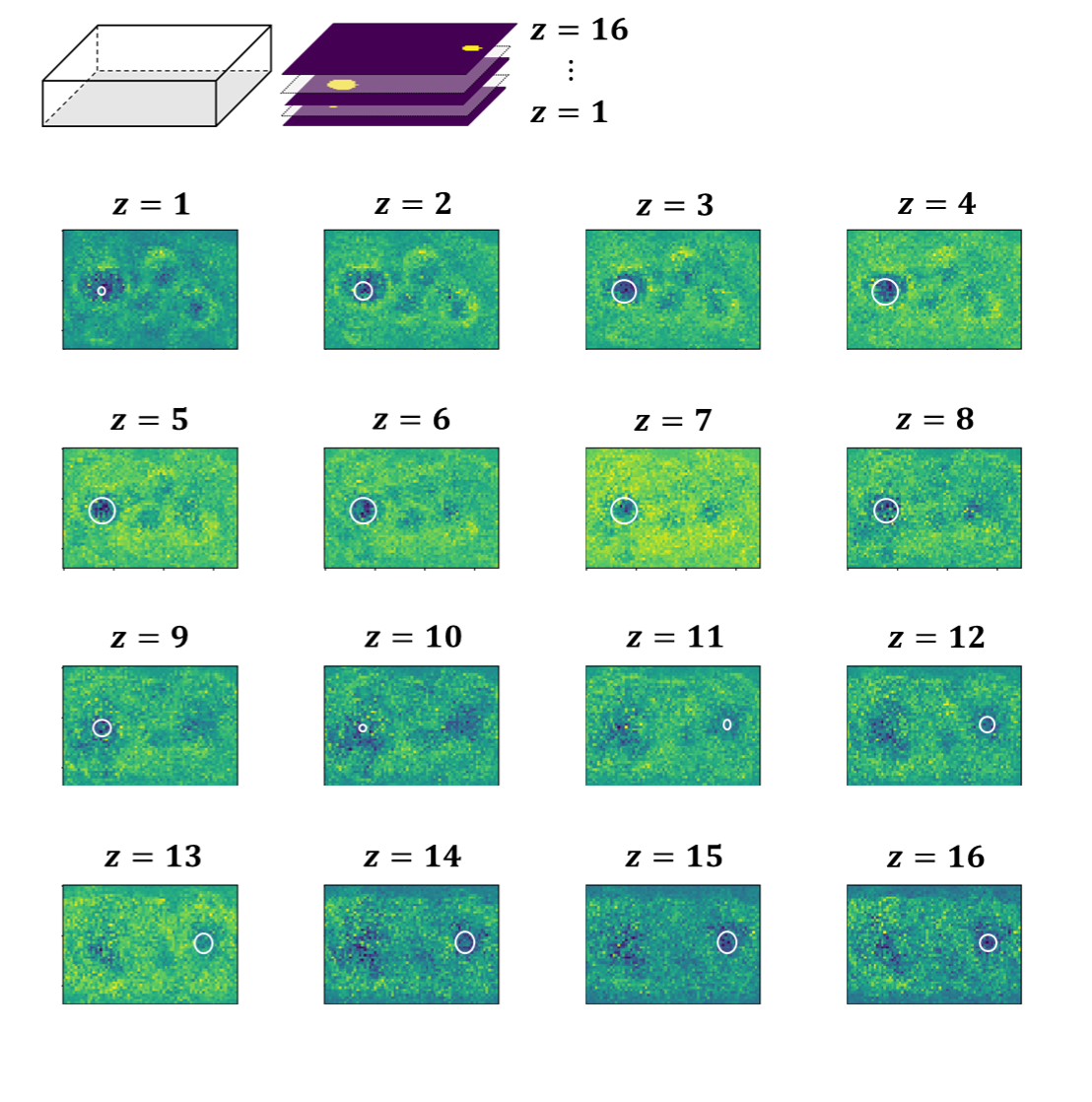}}
\vspace*{-0.5cm}
\centerline{\mbox{(a)}}
\centerline{\includegraphics[width=8cm]{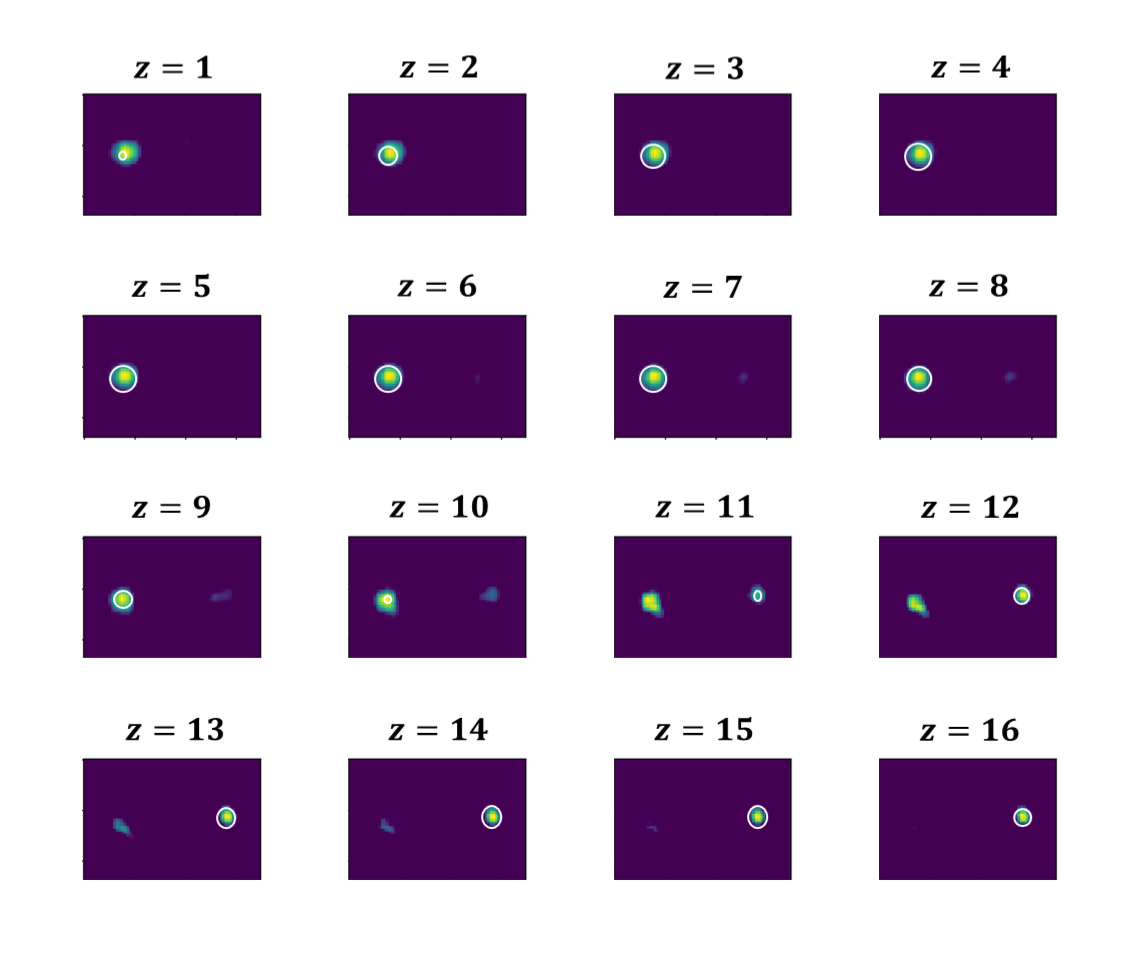}}
\vspace*{-0.5cm}
\centerline{\mbox{(b)}}
\caption{Slice-by-slice visualization of the feature map (a) after the fully connected layer, and (b) at the last layer.
The experiment was performed for the case of the numerical phantom in Fig. 3.
 }
\label{fig:comp}
\end{figure}

To show that the fully connected layer is working as a scaled version of appriximate inverse of the forward operator, we visualized the output after the fully connected layer. 
The results  in Fig.~\ref{fig:comp}(a) in comparison with the final results in Fig.~\ref{fig:comp}(b) reveal that the fully connected layer is indeed inverting the forward operator making a rough solution with a lot of noise. To generate  clean  final output, this results should be refined by the following convolution layers. As we have noted in the ablation study (Fig. 8), the consecutive convolution layers after the fully connected layer is very important and we need at least one single convolution layer with a single filter to recover the accurate location of the optical anomalies. 

\end{document}